\pdfoutput=1

\documentclass[11pt]{article}

\usepackage[]{acl}

\usepackage{times}
\usepackage{latexsym}

\usepackage{siunitx}
\usepackage[T1]{fontenc}

\usepackage[utf8]{inputenc}

\usepackage{microtype}
\usepackage{graphicx}
\usepackage{paralist}
\usepackage{multirow}
\usepackage[export]{adjustbox}
\usepackage{booktabs}
\usepackage{nicematrix}
\usepackage{xcolor}
\usepackage{colortbl}
\usepackage[outline]{contour}
\usepackage{enumitem}
\usepackage{siunitx}
\usepackage{subcaption}
\usepackage{comment}
\usepackage{amsmath}
\usepackage{amsfonts}
\usepackage{bbm}
\usepackage{booktabs}
\usepackage{siunitx}
\usepackage{makecell}
\usepackage{multirow}
\usepackage{hyperref}
\usepackage{inconsolata}

\definecolor{ETHPetrol}{RGB}{0,120,148}	%
\colorlet{MacroColor}{ETHPetrol}

\definecolor{ClementeColor}{RGB}{89,140,10}

\newcommand{\mymacro}[1]{#1}

\newcommand{\defn}[1]{\textbf{#1}}

\newcommand{\encoder}{\mymacro{\phi}}
\newcommand{\embdim}{\mymacro{D}}
\newcommand{\visembdim}{\mymacro{\embdim^{\prime}}}

\newcommand{\patches}{\mymacro{\Delta}}
\newcommand{\imgdim}{\mymacro{M}}
\newcommand{\imgdims}{\mymacro{\imgdim_1\times\imgdim_2}}
\newcommand{\strlen}{\mymacro{N}}
\newcommand{\channels}{\mymacro{C}}
\newcommand{\height}{\mymacro{H}}
\newcommand{\width}{\mymacro{W}}
\newcommand{\visencoder}{\mymacro{\psi}}
\newcommand{\flatener}{\mymacro{\textsc{flat}}}
\newcommand{\vlm}{\mymacro{\text{VLM}}}
\newcommand{\lm}{\mymacro{\text{LM}}}
\newcommand{\img}{\mymacro{x}}

\newcommand{\alphabet}{\mymacro{\Sigma}}

\newcommand{\connector}{\mymacro{\textsc{conn}}}
\newcommand{\connectorshort}{\mymacro{\textsc{c}}}

\newcommand{\nnor}{\mymacro{\mathcal{R}}}

\newcommand{\averageNnor}{\mymacro{\overline{\nnor}}}

\newcommand{\nn}{\mathcal{N}}

\newcommand{\condim}{\mymacro{\imgdim_C}}
\newcommand{\infoloss}{\mymacro{\mu}}

\newcommand{\defeq}{\mathrel{\stackrel{\textnormal{\tiny def}}{=}}}

\newcommand{\images}{\mymacro{I}}

\newcommand{\recon}{\mymacro{f_\theta}}

\newcommand{\loss}{\mymacro{\mathcal{L}}}

\newcommand{\str}{\mymacro{\sigma}}

\newcommand{\R}{\mymacro{\mathbb{R}}}

\newcommand{\N}{\mymacro{\mathbb{N}}}

\usepackage[textsize=scriptsize]{todonotes}
\title{Lost in Embeddings: Information Loss in Vision--Language Models}

\author{
    Wenyan Li$^{1}$\quad Raphael Tang$^{2}$\quad Chengzu Li$^{3}$\quad \textbf{Caiqi Zhang}$^{3}$ \\
     \quad \textbf{Ivan Vulić}$^{3}$ \quad \textbf{Anders Søgaard}$^{1}$ \\
   $^{1}$University of Copenhagen\quad $^{2}$Microsoft \quad 
    $^{3}$University of Cambridge \\
    {
    \{\href{mailto:weli@di.ku.dk}{\texttt{weli}},
    \href{mailto:soegaard@di.ku.dk}{\texttt{soegaard}}\}\texttt{@di.ku.dk}\quad
    \href{mailto:v-raptang@microsoft.com}{\texttt{v-raptang@microsoft.com}}}
    \\
    {
    \{\href{mailto:cl917@cam.ac.uk}{\texttt{cl917}},
    \href{mailto:cz391@cam.ac.uk}{\texttt{cz391}},
    \href{mailto:iv250@cam.ac.uk}{\texttt{iv250}}\}\texttt{@cam.ac.uk} 
}
}

\begin{document}
\maketitle

\begin{abstract}
    Vision--language models (VLMs) often process visual inputs through a pretrained vision encoder, followed by a projection into the language model's embedding space via a connector component. While crucial for modality fusion, the potential information loss induced by this projection step and its direct impact on model capabilities remain understudied. We introduce two complementary approaches to examine and quantify this loss by analyzing the latent representation space. First, we evaluate semantic information preservation by analyzing changes in $k$-nearest neighbor relationships between image representations, before and after projection. Second, we directly measure information loss by reconstructing visual embeddings from the projected representation, localizing loss at an image patch level. Experiments reveal that connectors substantially distort the local geometry of visual representations, with $k$-nearest neighbors diverging by 40–60\% post-projection, correlating with degradation in retrieval performance. The patch-level embedding reconstruction provides interpretable insights for model behavior on visually grounded question-answering tasks, finding that areas of high information loss reliably predict instances where models struggle.\footnote{Code: \url{https://github.com/lyan62/vlm-info-loss}.}
\end{abstract}

\section{Introduction}
\label{sec:intro}
Vision--language models (VLMs) have advanced on many tasks, e.g., visual question answering and image captioning by combining pretrained vision encoders with pre-trained language models. Many of these models employ a small neural network module, known as a connector (or a projector), to bridge the gap between the visual and textual representation spaces.
The connectors project visual representations into sequences of embeddings that a language model can process \citep{chen2024lion, liu2023llava, deitke2024molmo,laurençon2024matters, chen2024internvl,zhang2025mm, sun2024generative}.
Common connector architectures include multi-layer perceptrons (MLPs) or attention-based approaches \cite{jaegle2021perceiver,laurençon2024matters}.

\begin{figure*}[t]
    \centering
    \begin{subfigure}[b]{0.32\textwidth}
        \centering
        \includegraphics[width=\textwidth, trim={0pt 128pt 240pt 128pt}, clip]{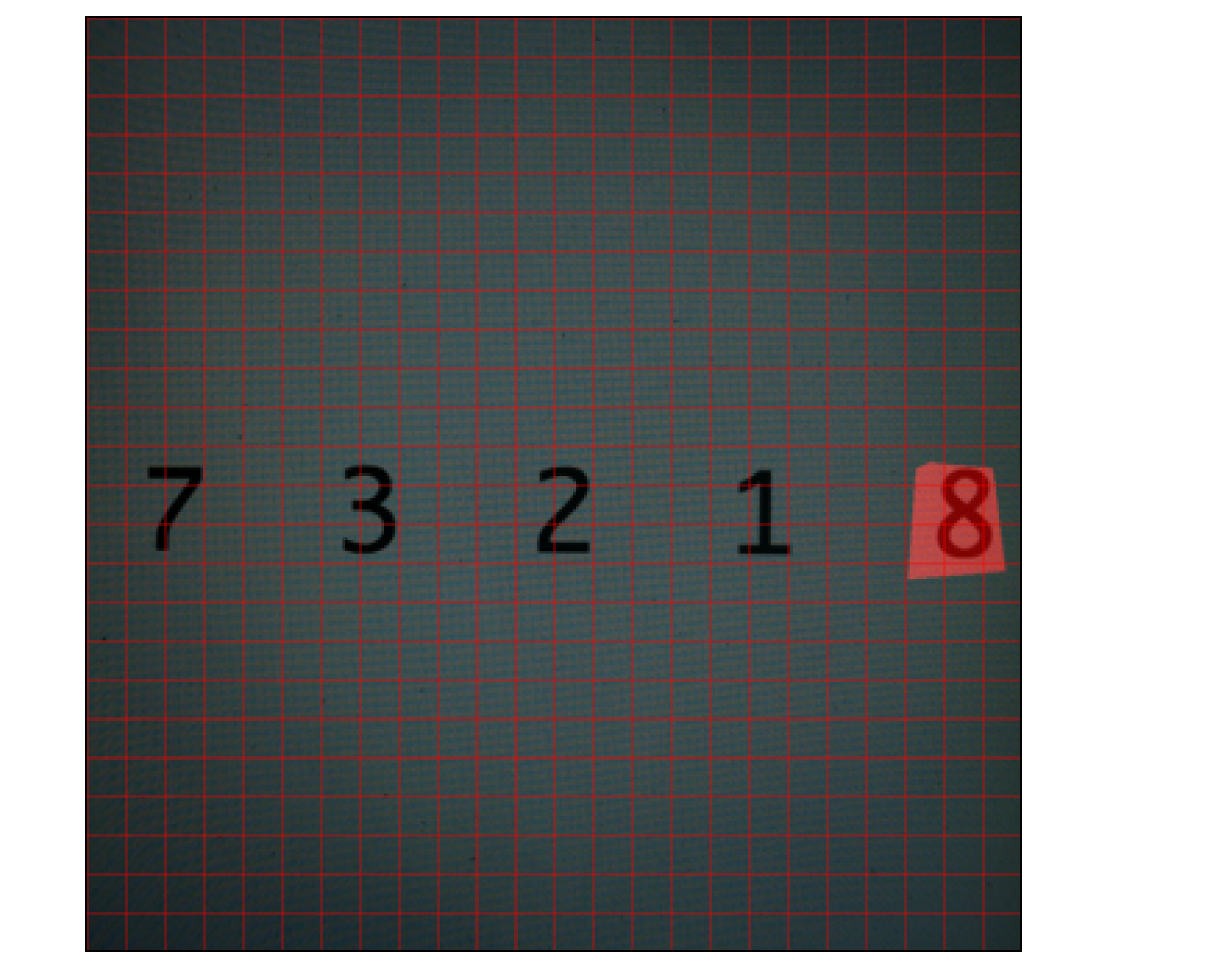}
        \caption{Input image with red answer mask}
        \label{fig:original}
    \end{subfigure}
    \hfill
    \begin{subfigure}[b]{0.32\textwidth}
        \centering
        \includegraphics[width=\textwidth, trim={40pt 128pt 200pt 128pt}, clip]{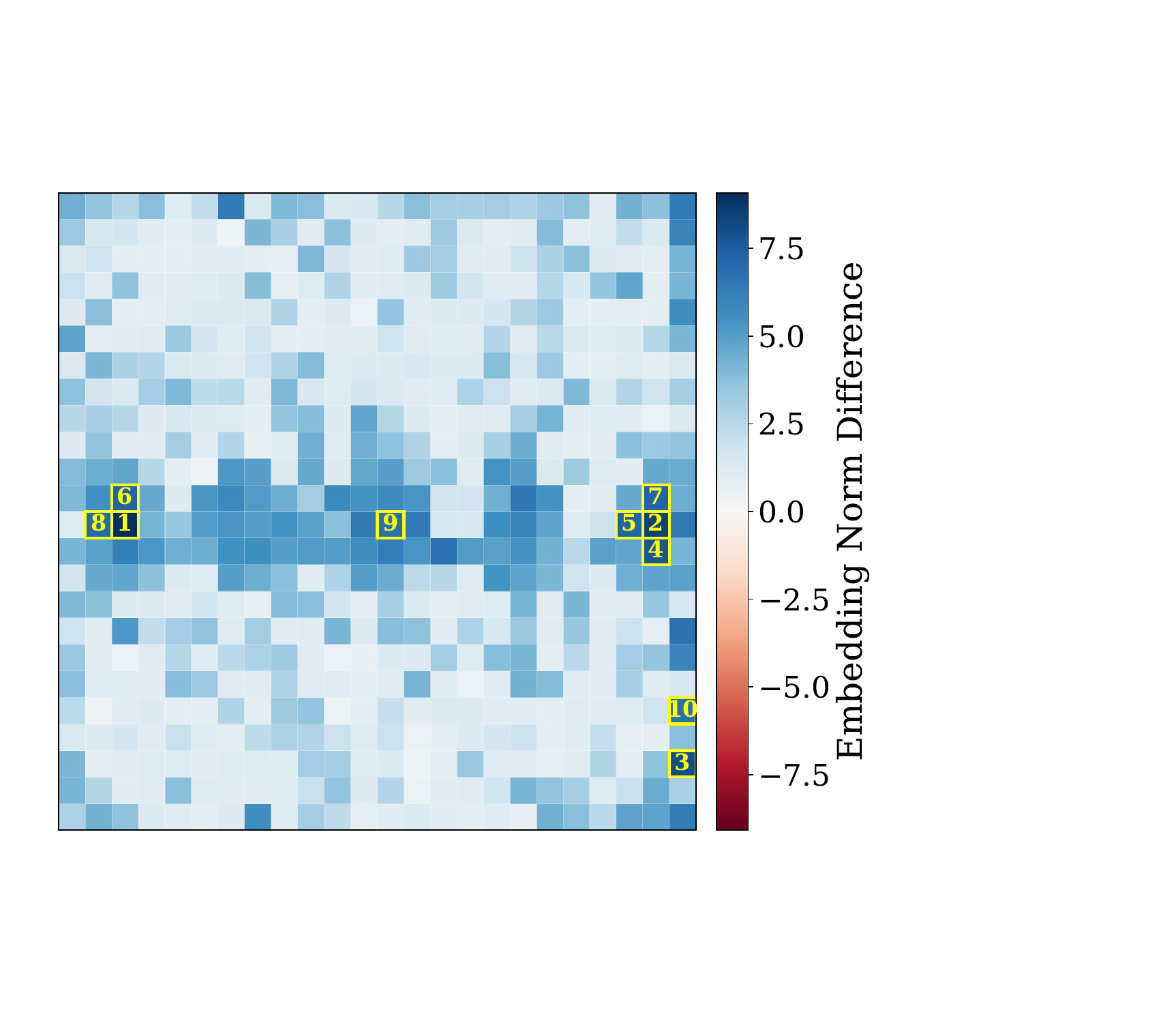}
        \caption{Embedding norm signed difference}
        \label{fig:heatmap}
    \end{subfigure}
    \hfill
    \begin{subfigure}[b]{0.32\textwidth}
        \centering
        \includegraphics[width=\textwidth, trim={40pt 128pt 200pt 128pt}, clip]{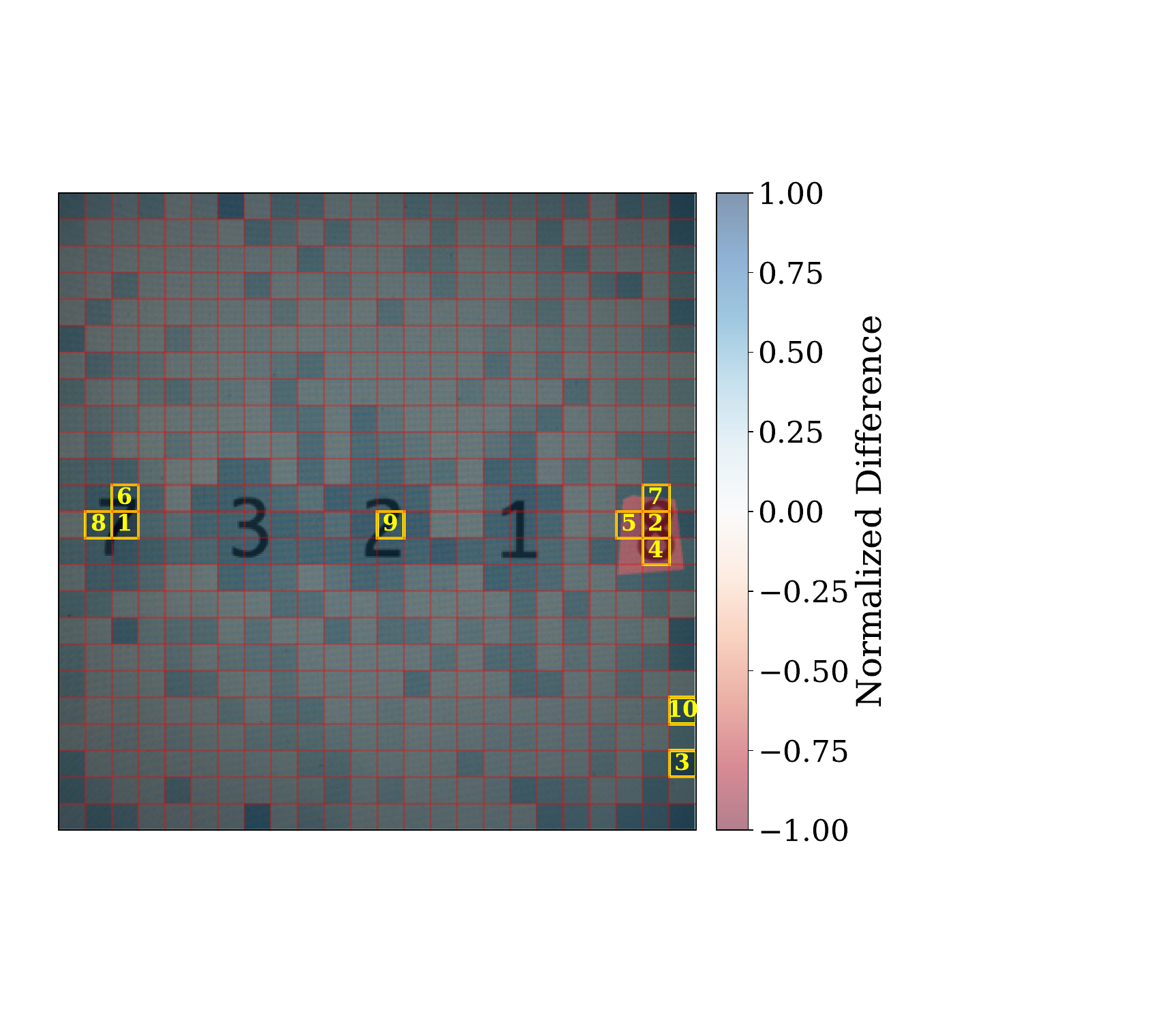}
        \caption{Image overlay with norm difference}
        \label{fig:overlay}
    \end{subfigure}
    \caption{Visualization of patch-wise information loss in the embeddings explains the incorrect predicted answer in VizWiz Grounding VQA. For the question ``What is the fifth number?'', LLaVA incorrectly predicted ``18''. Figure~\ref{fig:heatmap} display the difference between the $L^2$ norm of the original and the reconstructed patch embeddings. Blue regions indicate where original embeddings have larger norms than predicted embeddings, while red regions show where predicted embeddings have larger norms. The top 10 high-loss patches are marked by yellow squares. Figure~\ref{fig:overlay} shows high loss occurring in several answer-relevant patches contribute to the incorrect prediction.}
    \label{fig:patch_differences}
\end{figure*}

While these connector modules enable efficient cross-modal integration~\cite{li2024multimodal}, projecting rich visual features into embeddings compatible with language models typically involves dimensional conversion and representation restructuring. Naturally, this raises questions about potential information loss\footnote{In this paper, we use ``information loss" to broadly describe possible degradation of visual information, including aspects of the representation that cannot be recovered or directly observed after the projection. In a stricter sense, this could also be viewed as a representational gap or discrepancy rather than true loss, e.g. changes in the local geometry as reflected by k-nearest-neighbor relationships.} during projection, and how such loss impacts downstream task performance. As shown in Figure~\ref{fig:patch_differences}, the loss of critical visual details most relevant to answering the question imposes inherent limitations on the reasoning capabilities, since the language model's performance is constrained by the quality and completeness of the visual information it receives.

Despite the growing research on VLM connector architectures 
and their impact on downstream performance~
\cite{lin-etal-2024-preserve, zhu2025connectorssurveyconnectorsmultimodal}, there has been limited investigation into how well 
they preserve visual information in the latent space. Quantifying this information loss presents substantial challenges; 
traditional methods like canonical correlation analysis~\cite{cca} struggle with variable-length high-dimensional visual features processed through diverse connector architectures in vision-language models. Performance 
degradation can also take more than one form, adding to the complexity of its study. For instance, it can take the form of a \emph{direct information loss} due to an inherently lossy connector, or a \emph{geometric collapse} where distinct features become entangled in the projected embedding space.

To bridge this gap in the literature, we present an evaluation framework to quantify information loss in VLM connectors from both the geometric perspective and that of localized information loss. We first examine if the connector projection changes the geometric structure of latent visual representations. By introducing \textbf{$k$-nearest neighbors overlap ratio}, we measure how much the neighborhoods of image embeddings change before and after the projection in the latent representation space, thereby estimating how well geometric and semantic relationships are preserved.\footnote{Here, the semantic relationship denotes the semantic similarity between pairwise embeddings measured by $L^2$ distance or their inner product similarity.} Second, we quantify localized information loss by training a model to reconstruct the original visual embeddings from the projected embeddings. This \textbf{patch-level visual embedding reconstruction} allows us to pinpoint the high-loss regions in the image---areas where visual features are hard to recover after projection (Figure~\ref{fig:patch_differences}). This two-step approach provides both quantitative analysis and interpretable visualizations, offering insights into the nature of information transformation during vision-text integration.

\section{VLMs and Connectors}
\label{sec:background}

Integrating visual and textual inputs is fundamental for VLMs to process multimodal information effectively. 
Existing VLMs typically employ two main approaches \cite{li2024multimodal}: models like LLama3.2~\cite{grattafiori2024llama3herdmodels} and BLIP~\cite{pmlr-v202-li23q} leverage cross-modal attention mechanisms, while others such as LLaVA~\cite{liu2023llava} and Qwen-2-VL~\cite{bai2025qwen25vltechnicalreport} adopt connectors to project visual representations into latent vectors compatible with large language models (LLMs).\footnote{In this paper, we do not consider VQ-VAE~\cite{vqvae} based VLMs, which are more often used for text-to-image generation.}

\citet{lin-etal-2024-preserve} categorize connectors into two types: feature-preserving and feature-compressing connectors. Feature-preserving connectors include MLPs that preserve the number of patch embeddings, such as the two-layer MLP connector in LLaVA. In contrast, feature compressing connectors project image patch embeddings to a shorter sequence, often involving transformer-based or convolution architectures with pooling operations over the original vision embedding \citep{jaegle2021perceiver}. Feature compressing connectors include the perceiver sampler in Idefics2~\cite{laurençon2024matters} and the patch merger in Qwen-2-VL~\cite{bai2025qwen25vltechnicalreport}. In this paper, we estimate information loss in both types of connectors.
    
\subsection{Formalizing Encoders and Connectors}
\label{subsec:definition}

We now give a formal definition of connectors.
First, we consider the textual input. 
Let $\alphabet$ be an alphabet of symbols. 
A \defn{string encoder}, $\encoder$, is a function that maps a string $\str$ to a sequence of real-valued representations. 
Formally,
\begin{equation}
\encoder \colon \alphabet^\strlen \rightarrow (\R^{\embdim})^\strlen,
\end{equation}
where $\strlen \in \N$ is a parameter in the dependent type that denotes the length of the input string, and $\embdim$ is the dimensionality of the representation.
Next, we turn to the visual input. 
Let $\patches= \{1, \ldots, 256\}^{\height \cdot \width\times\channels}$ be an array of image patches, where $\height$ and $\width$ represent the height and width dimensions, and $\channels$ is the number of color channels per pixel. 
A two-dimensional image of patch dimensions $\imgdims$ can thus be represented as an element of $\patches^{\imgdims}$. Where $\patches^{\imgdims}$ denotes the set of all possible $\imgdims$ grids of patches.
The \textbf{vision encoder} is formalized as a dependent type: 
\begin{equation}
\visencoder \colon \patches^{\imgdims} \rightarrow (\R^{\visembdim})^{\imgdims},
\end{equation}
where $\imgdim_1$ and $\imgdim_2$ are parameters in the dependent type, representing the grid dimensions of the image patches, and $\embdim'$ is the visual embedding dimension. This maps a grid of image patches to a grid of embedding vectors.

A \defn{connector} module transforms the vision encoder's output to match the dimensionality of the text encoder---projecting visual embeddings of dimension $\visembdim$ to text-compatible dimension $\embdim$. We define the connector as a function of type:
\begin{equation}
\label{eq:conn}
    \connector \colon (\R^{\visembdim})^{\imgdims} \rightarrow (\R^{\embdim})^{\condim},
\end{equation}
where we typically have $\condim \le \imgdim_1\imgdim_2$. We also use $\connectorshort$ as shorthand for $\connector$.

For combining the output of the string encoder and the vision encoder, we define a \defn{flattener} that combines visual and textual embeddings into a unified sequence:
\begin{equation}
\label{eq:flat}
    \flatener \colon (\R^{\embdim})^{\condim} \times (\R^{\embdim})^\strlen \rightarrow (\R^{\embdim})^{\condim + \strlen}
\end{equation}
This creates a sequence of length $\condim + \strlen$ by concatenating the flattened grid of visual embeddings with the sequence of text embeddings.

The complete vision--language models we consider can then be expressed as the a composition of these functions:
\begin{equation}
\vlm(\img, \str) = \lm(\flatener(\connector(\visencoder(\img)), \encoder(\str)))
\end{equation}
where $\img \in \patches^{\imgdims}$ is an input image, $\str \in \alphabet^\strlen$ is an input text sequence, and $\lm$ is an auto-regressive language model that predicts probability of next tokens.

We focus on quantifying the information loss at the connector module defined in Equation~\ref{eq:conn}. Formally, the information loss over the connector is a function $\infoloss: (\visencoder(\img), \connector(\visencoder(\img))) \to \R_{\ge 0}$. We explore how such loss correlate and explain model performance.

\section{Quantifying Information Loss}
\label{sec:method}

 \begin{figure}[t]
    \centering
        \includegraphics[width=\columnwidth, trim={0 0 0 0.5em}, clip]{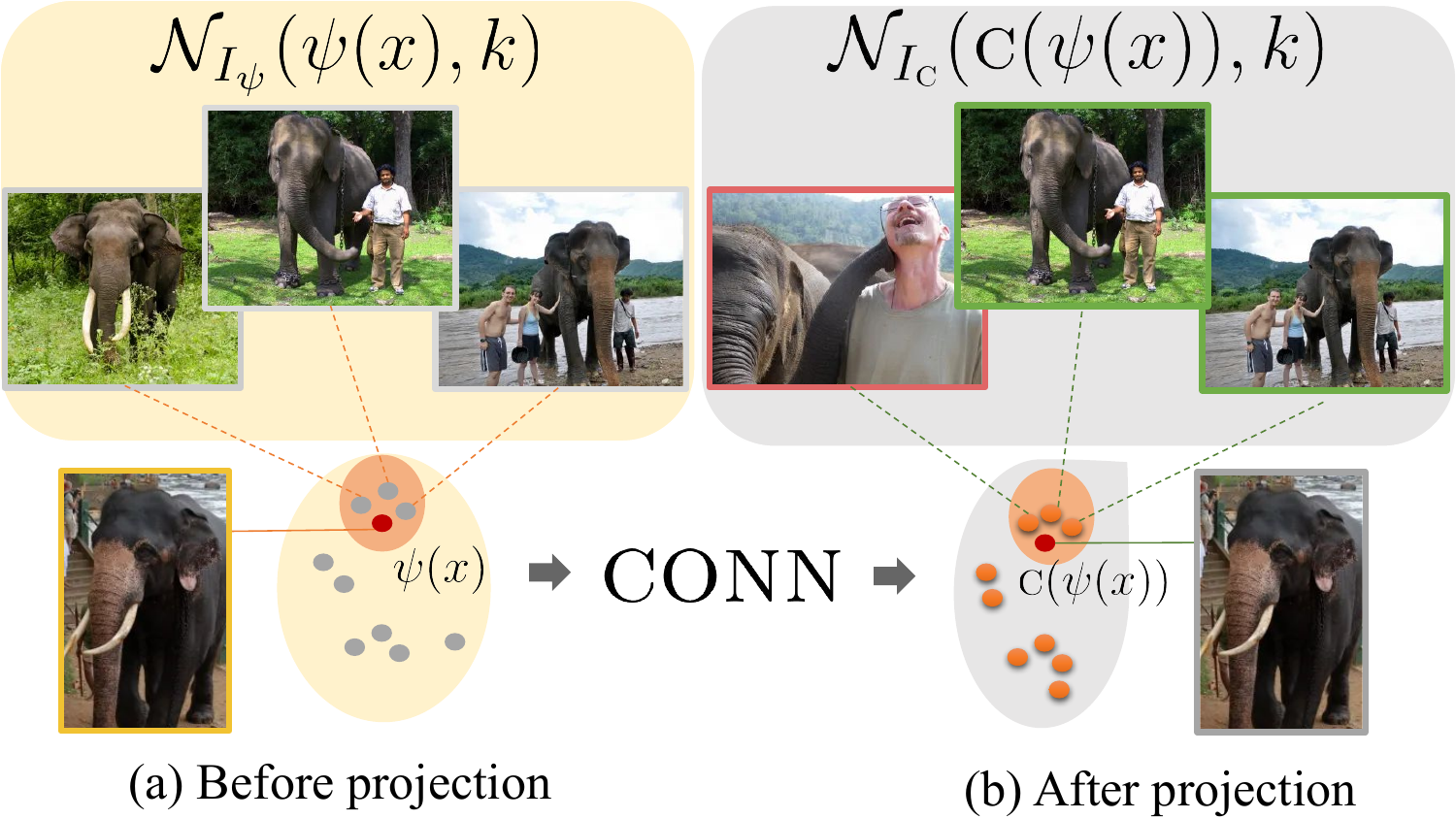}
        \caption{The $k$-nearest neighbors overlap ratio measures the overlap of an image's neighbors before and after projection. In this example, with $k=3$, the overlap ratio is 0.67 because two out of the three nearest neighbors are identical in both representation spaces.}
        \label{fig:neighbors}
\end{figure}

We propose two methods for quantifying information loss over the projection step described above. The first method quantifies the preservation of structural information in semantic embeddings by comparing each image representation's $k$-Nearest Neighbors ($k$-NN, \citet{fix1951discriminatory}) before and after projection. The overlap ratio of the $k$-NN neighbors captures how well local geometry of the semantic embeddings are preserved in the latent space. Figure~\ref{fig:neighbors} gives an example where two of the three nearest neighbors overlap before and after projection.
The second method evaluates patch-level representation (Figure~\ref{fig:patch_differences}) distortion by training an \emph{ad hoc} neural network to reconstruct the original image embedding from its projected representation, detailed in Section~\ref{subsec:embed-reconstruct}.

\subsection{$k$-Nearest Neighbors Overlap Ratio}
\label{subsec:neighborhood-overlap}

To quantify geometric information loss during projection in visual representation spaces, we propose the \textbf{$k$-nearest neighbors overlap ratio (\textsc{KNOR})}, which measures how well $k$-NN relationships among image embeddings are preserved before and after projection through the connector.

Let $\images$ be a finite set of images, $\visencoder$ a vision encoder, and $\connector$ ($\connectorshort$ for short) a connector as described in \S\ref{subsec:definition}. We use $\images_\visencoder=\{\visencoder(\img)\}_{\img\in\images}$ to indicate the family of embedded images, and $\images_\connectorshort=\{\connector(\visencoder(\img))\}_{\img\in\images}$ for the projection of the embedded images.
The \text{$k$-NN} overlap ratio for an image $\img$ is defined as
\begin{equation}
    \nnor(\img, k) \defeq \frac{ \big| \nn_{\images_\visencoder}(\visencoder(\img), k) \cap \nn_{\images_\connectorshort}(\connectorshort(\visencoder(\img)), k) \big| }{ k }
    \label{eq:overlap-ratio}
\end{equation}
Where $\nn_{\images_\visencoder}(\visencoder(\img), k)$ is the set of $k$-nearest neighbors of $\visencoder(\img)$ among the pre-projected embeddings, and $\nn_{\images_\connectorshort}(\connectorshort(\visencoder(\img)), k)$ is the set of $k$-nearest neighbors of $\connectorshort(\visencoder(\img))$ among the projected embeddings. The \defn{average overlap ratio} is given by
\begin{equation}
    \averageNnor(k) \defeq \frac{1}{|\images|} \sum_{\img\in\images} \nnor(\img, k)
\end{equation}

The average overlap ratio measures how well the local geometric structure is preserved after projection. An optimal connector would maintain the same $k$-NN sets for $\visencoder(\img)$ and $\connectorshort(\visencoder(\img))$. Lower overlap ratio corresponds to more geometric information loss due to projection, while higher overlap suggests faithful geometric retention.

\subsection{Embedding Reconstruction}
\label{subsec:embed-reconstruct}

\label{subsec:reconstruct}
\textsc{KNOR} reflects structural information loss during projection, indicating how well local geometric relationships among image embeddings are preserved. However, it does not identify the loss of fine-grained visual features at the patch level.

To address this, we further quantify and localize patch-level information loss by attempting to reconstruct the original vision embeddings from their projected representations.

Specifically, given a connector $\connector$ defined in Equation~\ref{eq:conn}
and set of images $\images\subset \patches^{\imgdims}$, we train a \defn{reconstruction model} 
$\recon: (\R^{\embdim})^{\condim} \rightarrow (\R^{\visembdim})^{\imgdims}$ to minimize reconstruction loss.
For each patch index $(i, j)\in \imgdims$, we define the per-patch loss as 
\begin{equation}
\loss_{\text{patch}}(\img, i, j) \defeq \|\visencoder(\img)_{(i,j)} - \recon(\connectorshort(\visencoder(\img)))_{(i,j)}\|_2^2
\label{eq:reconstruction}
\end{equation}
which measures the squared Euclidean distance between the original vision embedding and its reconstruction for each patch.
The total reconstruction loss is therefore the sum of the patch-wise losses across all patches and images:
\begin{equation}
\loss_{\text{recon}}(\images) \defeq \sum_{\img\in\images} \sum_{\substack{(i,j) \in \\  \imgdims}} \loss_{\text{patch}}(\img, i, j)
\end{equation}

This patch-wise reconstruction enables us to identify and visualize the spatial distribution of information loss across the image.

\section{Experimental Setup}
\label{sec:setup}
    We quantify information loss using both methods across three open-weights connector-based vision-language models on six datasets spanning question answering, captioning, and retrieval tasks. We assume that greater structural and semantic information loss during projection through the connector leads to reduced neighborhood overlap, while greater patch-wise information loss results in higher reconstruction error.

    \subsection{Pretrained VLMs}
    \label{subsec:models}
    We consider three open-weights connector-based vision-language models including LLaVA~\cite{liu2023llava}, Idefics2~\cite{laurençon2024matters}, and Qwen2.5-VL~\cite{bai2025qwen25vltechnicalreport}. 
    LLaVA uses a two-layer MLP as the connector, preserving total number of patches for each image. In contrast, Idefics2 uses an attention-based perceiver resampler \cite{jaegle2021perceiver} that projects image embeddings to a fixed-length sequence of embeddings. Qwen2.5-VL uses a MLP-based patch merger which merges every four neighboring patch representations into one. We use the 7B-instruct model variants for LLaVA and Qwen2.5-VL, and the Idefics2-8B-instruct model. 
    
    \subsection{Evaluation Datasets}
    \label{subsec:dataset}
        We evaluate on six diverse datasets, each of which probes different aspects of visual understanding. 
        \begin{description}[nosep, leftmargin=1em] %
        \item \textbf{SEED-Bench} \cite{Li_2024_CVPR} provides categorized multiple-choice questions spanning cognitive tasks from basic scene understanding to complex visual reasoning.
        \item \textbf{VizWiz Grounding VQA} \cite{chen2022grounding} includes images taken by blind or low-vision individuals regarding scenarios that require visually-grounded question answering.
        \item \textbf{VQAv2}~\cite{VQA} covers open-ended questions that test general visual comprehension.
        \item \textbf{CUB-200-2011} \cite{CUB} is a commonly used dataset for fine-grained image retrieval that covers 200 species of birds.
        \item \textbf{Flickr30k} \cite{flickr30k} and \textbf{COCO} \citep{lin2014coco} Karpathy test set \citep{cocokarpathy} are used for image captioning evaluation.
    \end{description}
    Together, these datasets offer complementary perspectives on how different types of visual information are preserved during projection and how information loss impacts various downstream tasks.

    \begin{figure*}[!t]
        \centering
        \includegraphics[width=0.86\linewidth, trim=0em 1em 0em 1em, clip]{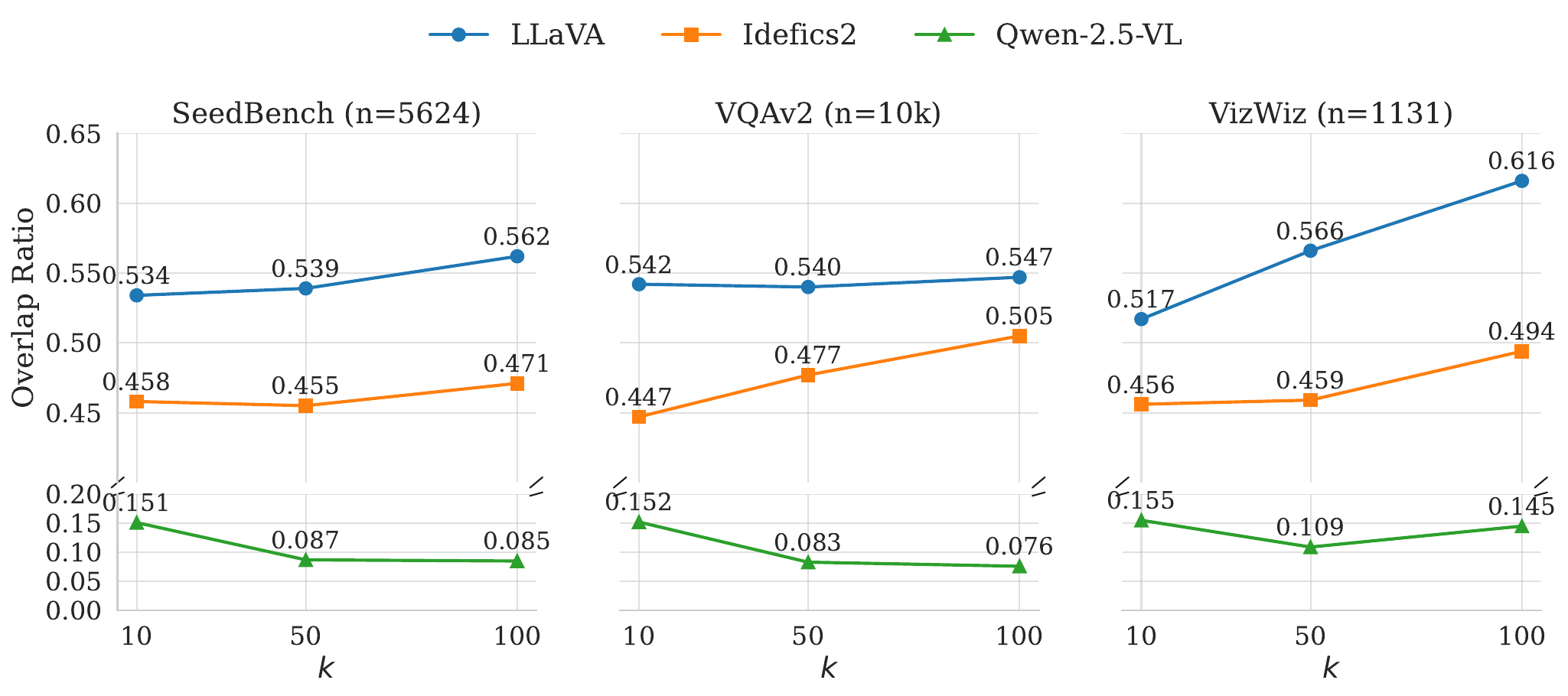}
        \caption{Neighborhood overlap ratios across three datasets: SeedBench validation, a 10,000-sample subset of VQAv2 validation, and Vizwiz grounding VQA validation. Analysis using 10, 50, and 100 nearest neighbors shows overlap ratios below 0.62 for all models, suggesting connectors poorly preserve geometric relationships and neighbor rankings for the visual representations.}
        \label{fig:neighborhood-overlap}
    \end{figure*}

    \subsection{Embedding Reconstruction Models}
    \label{subsec:reconstruction}

    We build models to reconstruct image patch embeddings from connector outputs. These reconstruction models are intentionally designed with larger capacity than the original connectors, including expanded hidden dimensions and additional hidden layers. This controlled setup ensures our models are trained to recover the original visual representations without creating new bottlenecks in the reconstruction process.

   \paragraph*{Architecture} We tailor our reconstruction models to each VLM's connector architecture. For LLaVA, which preserves the number of image patches during projection, we use a simple three-layer MLP with a 2048-dimension hidden layer. For Idefics2 and Qwen2.5-VL, which compress sequence length from $\imgdims$ to $\condim$, we implement transformer-based models to handle the differences in sequence length. The reconstruction model projects connector outputs to hidden embeddings with positional encodings before processing them through a 16-layer, 16-head transformer encoder with 2048-dimensional vectors. Table~\ref{tab:model_params} summarizes the parameters of the reconstruction models and their input and output dimensions.

    \begin{table}[tb]
        \centering
        \resizebox{\columnwidth}{!}{%
        \begin{tabular}{l r r r r}
        \toprule
        \textbf{Model} & $ \imgdim_1\imgdim_2 \times \visembdim $ &  $ \condim \times \embdim $ & \textbf{$|\connector|$} & \textbf{$|\recon|$} \\
        \midrule
        LLaVA & $576\times1024$ & $576\times4096$ & 21M & 27M \\
        Idefics2 & $576\times1152$ & $64\times4096$ & 743M & 844M \\
        Qwen2.5-VL & $576\times1280$ & $144\times3584$ & 45M & 843M \\
        \bottomrule
        \end{tabular}%
        }
        \caption{Model parameters and embedding dimensions. $|\connector|$ denotes number of parameters in the connector and $|\recon|$ represents number of parameters of the reconstruction model. Pre- and post-projection embedding dimensions are listed as $ \imgdim_1\imgdim_2 \times \visembdim $ and $\condim \times \embdim $.}
        \label{tab:model_params}
    \end{table}

    \paragraph*{Training} \label{para:recon-train}
    We train each of the embedding reconstruction models on the COCO 2017 train set~\cite{lin2014coco} for 30 epochs with early stopping. We apply a learning rate of $\num{1e-4}$, dropout of $0.1$, and a total batch size of $128$. For training stability, we apply normalization to both pre- and post-projection embeddings using mean and standard deviation of the dataset.

\section{Neighbor Rankings and Structural Information are Not Preserved}
\label{sec:res-neighborhood}

    We calculate \textsc{KNOR} (Section~\ref{subsec:neighborhood-overlap}) for images in the SeedBench validation set, a subset of the VQAv2 validation set with  $10,000$ images, and the validation set of Vizwiz grounding VQA dataset. It is intuitive that higher neighborhood overlap ratios suggest that the projection better preserves the structural relationships between image embeddings. As the neighborhood rankings directly impact image retrieval tasks, we also evaluate retrieval performance on the CUB dataset using both pre- and post-connector visual embeddings.
    
    \subsection{Low Overlap Ratio for All Models}
    \label{para:res-ratio}
    
    In Figure~\ref{fig:neighborhood-overlap}, we show the neighborhood overlap ratio across $k=10$, 50, and 100 nearest neighbors, averaging through all unique images in the evaluation datasets.\footnote{Visual embeddings pre- and post-connector projection have a 1-1 mapping to the input image, and these visual embeddings are not impacted by the language model prompts.} The neighborhood overlap ratios for LLaVA and Idefics-2 are around 50\%. LLaVA achieves its highest overlap of 61.6\% at $k=100$, while Qwen2.5-VL loses nearly 90\% of the neighborhood ranking information. This suggests a significant reordering of nearest neighbors post-projection across all models. While LLaVA maintains higher structural preservation compared to Qwen2.5-VL and Idefics-2, it shows notable neighbor reshuffling, especially at smaller neighborhood sizes (k=10). 
    
\begin{figure}[!t]
        \centering
    
        \begin{subfigure}[b]{\columnwidth}
            \centering
            \includegraphics[width=0.95\columnwidth]{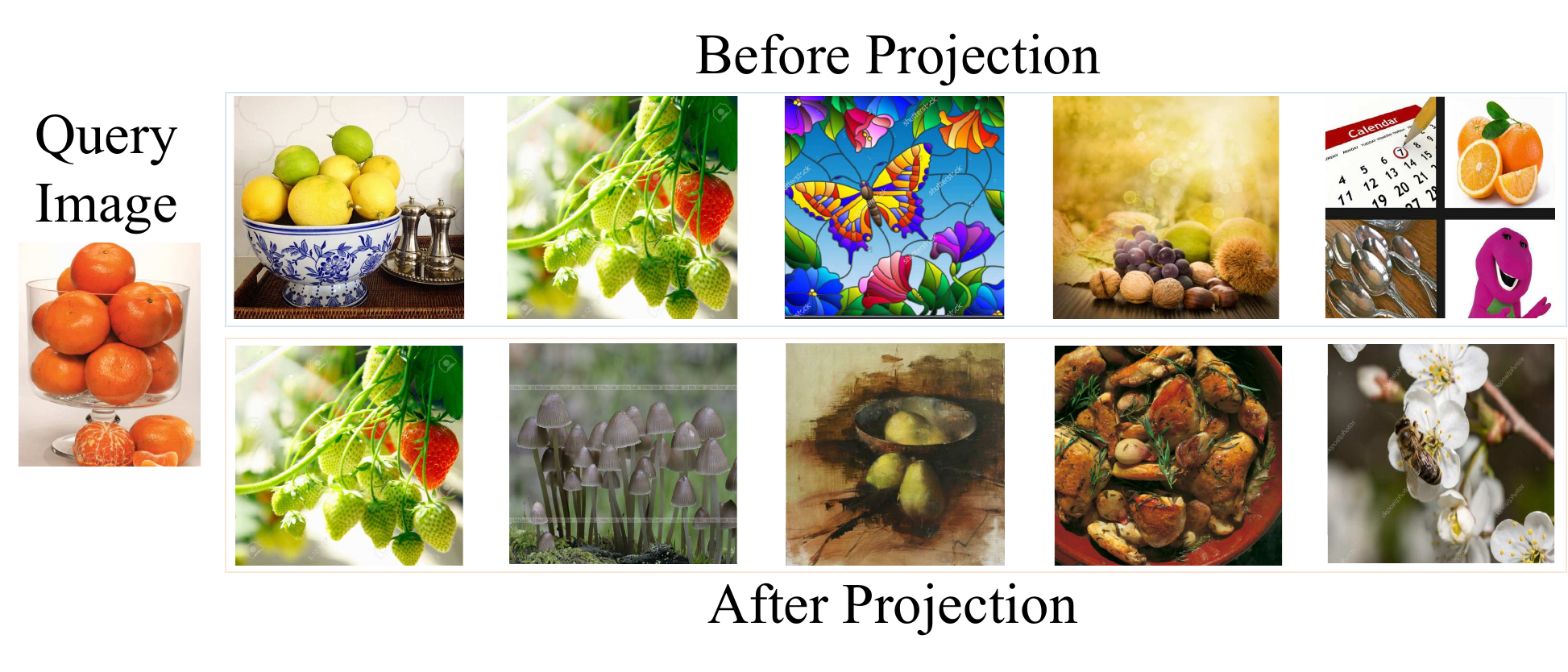}
            \caption{Five nearest neighbors of LLaVA image embeddings}
        \end{subfigure}

        \begin{subfigure}[b]{\columnwidth}
            \centering
            \includegraphics[width=0.95\columnwidth]{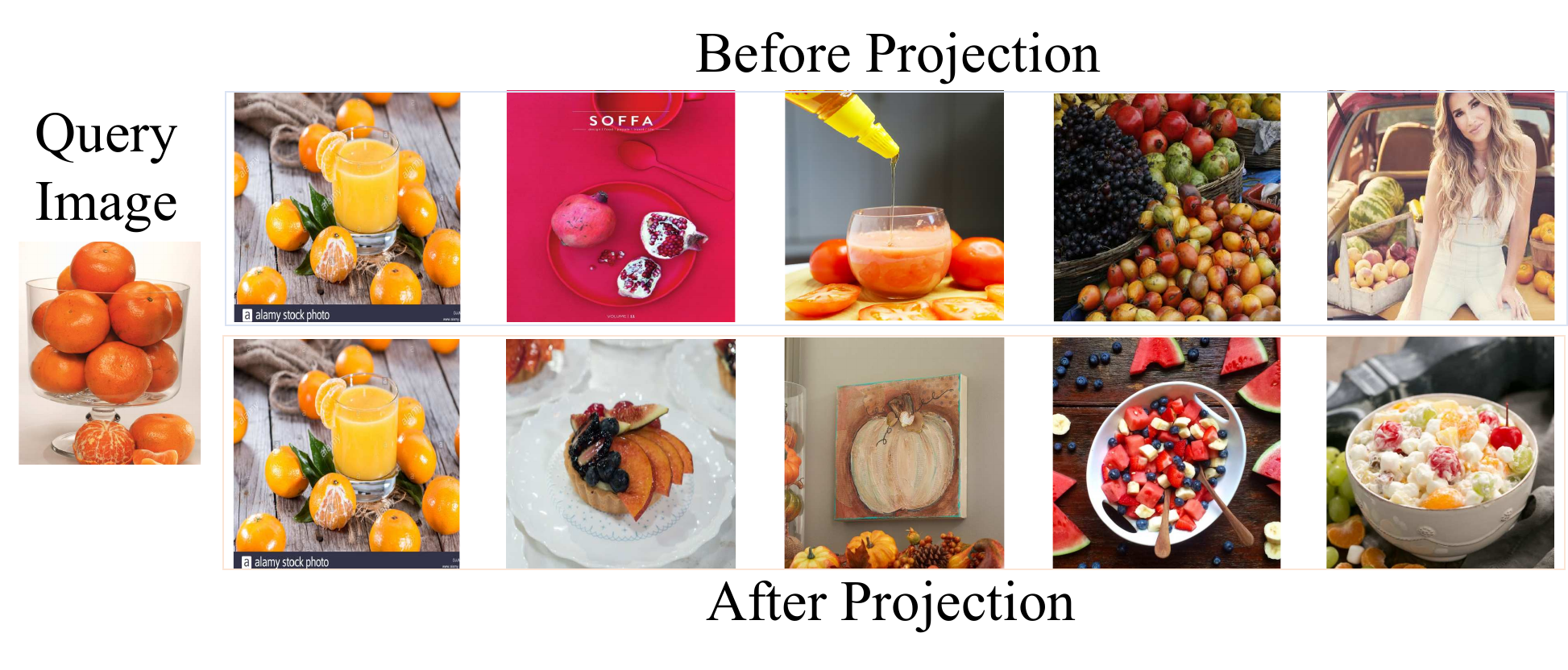}
            \caption{Five nearest neighbors of Idefics2 image embeddings}
        \end{subfigure}

        \begin{subfigure}[b]{\columnwidth}
            \centering
            \includegraphics[width=0.95\columnwidth]{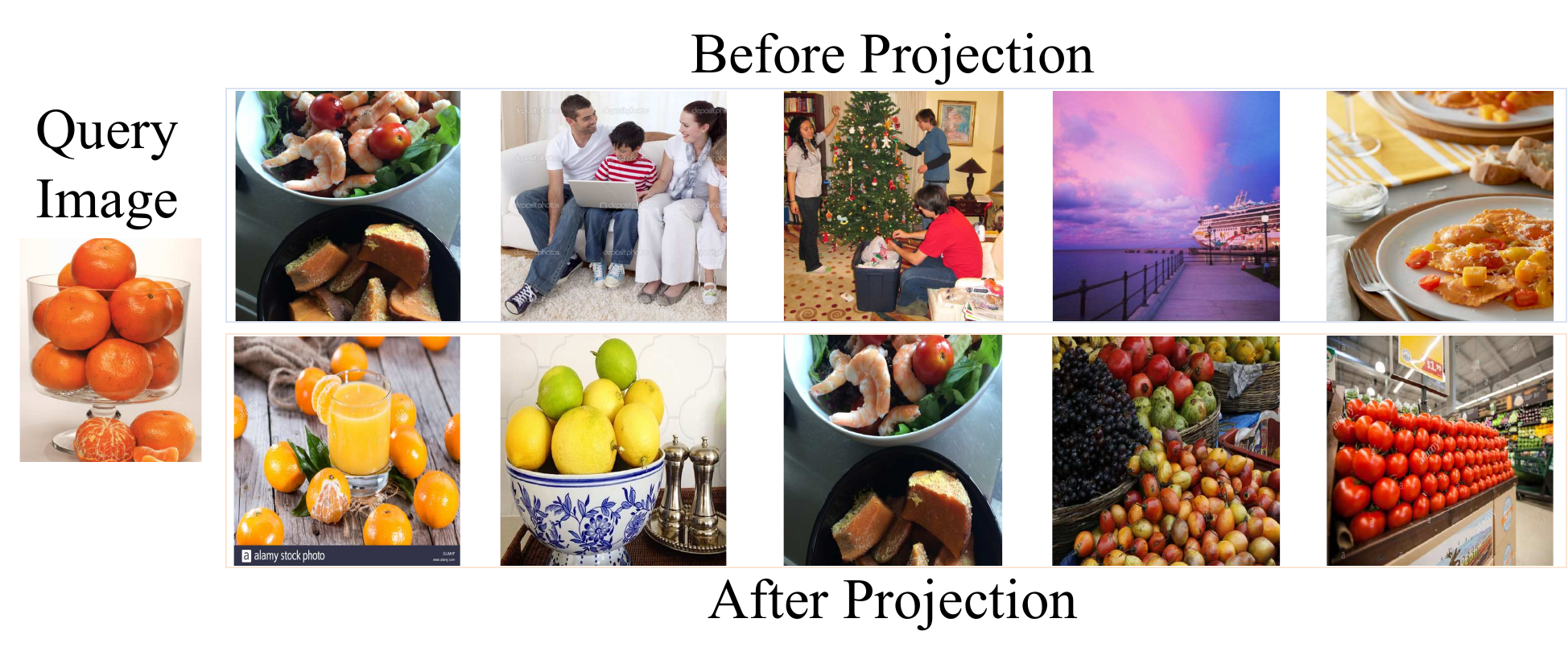}
            \caption{Five nearest neighbors of Qwen2.5-VL image embeddings}
        \end{subfigure}
    
        \caption{Comparison of five nearest neighbors searched with pre-projection (top) and post-projection (bottom) embeddings using different models. The first image in each row is the query image, followed by its nearest neighbors. For Qwen2.5-VL, despite a low neighborhood overlap ratio, post-projection embeddings retrieve more semantically similar images.}
        \label{fig:overlap-figs}
    \end{figure}

    In Figure~\ref{fig:overlap-figs}, we visualize the nearest neighbors of a given query image, revealing significant neighbor reordering across all models. However, for Qwen2.5-VL, the neighbors obtained with post-projection embeddings are more semantically similar to the query image. We suspect that this phenomenon could stem from its continuous training of the image encoder in the pretraining stage and the patch merging, which yields more semantically meaningful post-projection embeddings. Other VLMs such as LLaVA use a frozen vision encoder, where the connector is updated to inherit features from the pretrained encoder. However, in Qwen2.5-VL, continued pretraining with an unfrozen vision encoder produces fundamentally different learned visual embeddings. This indicates that the pre- and post-projection visual representations are not equivalent, but may not necessarily lead to worse semantic representations of the image.

\begin{table}[tb]
    \centering
    \adjustbox{max width=\columnwidth}{
    \begin{tabular}{l l cc |rr}
    \toprule
    \textbf{Model} & \textbf{Emb} & \textbf{$\bar{R}$} & \textbf{$\rho$} & \multicolumn{2}{c}{\textbf{Recall}} \\
    \cmidrule(lr){5-6}
    & & & &  \textbf{R@1} & \textbf{R@5} \\
    \midrule
    \multirow{2}{*}{LLaVA} 
    & Pre & \multirow{2}{*}{0.40} & 0.08 & 8.34 & 21.82 \\
    & Post & & 0.11 & 6.16 & 17.22 \\
    \midrule
    \multirow{2}{*}{Idefics2}
    & Pre & \multirow{2}{*}{0.39} & 0.23 & \textbf{13.10} & \textbf{30.81} \\
    & Post & & \textbf{0.28} & 10.87 & 25.28 \\
    \midrule
    \multirow{2}{*}{Qwen-2.5-VL}
    & Pre & \multirow{2}{*}{0.08} & 0.10 & 4.23 & 11.74 \\
    & Post & & 0.11 & 10.65 & 26.44 \\
    \bottomrule
    \end{tabular}
    }
    \caption{Zero-shot retrieval performance on the CUB test set using $L^2$ distance as the similarity measure. R@$k$ denotes Recall at rank $k$. We also report average overlap ratio $\bar{R}$. The Spearman correlation coefficient $\rho$ is calculated between R@$5$ and $k$-nearest neighbor overlap ratio for each sample, with $k=100$. All correlation scores are statistically significant with $p < \num{1e-3}$.
    \label{tab:cub_correlation}}
\end{table}

    \subsection{Image Retrieval Evaluation}
    \label{subsec:img-ret}
    
    To verify if neighborhood reordering correlates with a degradation in the semantic representation of images, we evaluate on the CUB-200-2011 image retrieval test set~\cite{CUB}. We perform zero-shot image retrieval with pre- and post-connector embeddings for each query image, excluding the query image itself from the gallery. The pre-and post-projection embeddings are indexed with FAISS~\cite{douze2024faiss}, and we experiment with retrieving similar images based on both the $L^2$ distance and the inner product similarity (Table~\ref{tab:cub_res} in Appendix) of the image representations.

    We report Recall@1 (R@1) and Recall@5 (R@5) in Table~\ref{tab:cub_correlation}. Consistent with the neighborhood overlap visualization in Figure~\ref{fig:overlap-figs}, we observe degradation in R@5 of 41.4\% for LLaVA and 18.8\% for Idefics2 when using post-projection image embeddings for retrieval. In contrast, Qwen2.5-VL shows improved retrieval performance with post-projection embeddings, suggesting that its low overlap ratio reflects substantial differences between pre- and post-projection representations.

    To examine how structural preservation relates to retrieval, we compute Recall@5 and the 100-nearest-neighbor overlap ratio (KNOR) for each sample, then calculate their Spearman correlation. As shown in Table~\ref{tab:cub_correlation}, all models show a positive correlation, with coefficients of about 0.1 for Qwen2.5-VL and LLaVA, and a stronger correlation of 0.3 for Idefics2. All p-values are below $1e-3$, indicating statistical significance. This positive per-sample correlation means that, within a given model, images whose local neighborhoods are better preserved tend to achieve higher retrieval performance. For Qwen2.5-VL, we observe a small but positive per-sample correlation between local overlap ratio and recall score. This suggests that while most of the pre-projection structure was discarded to create a more semantically meaningful space, retaining certain stable neighborhoods remains advantageous for specific images. Retrieval examples are shown in Figure~\ref{fig:cub} in the Appendix.

\section{Reconstruction and Model Behavior} 
\label{sec:reconstruction-res}

    While the neighborhood overlap ratio reflects structural information loss in the semantic representation space, we further examine the information loss at the image patch level. Specifically, as in Equation~\ref{eq:reconstruction}, we reconstruct patch-level visual representation $\visencoder(\img)$ of an image from its projected counterpart $\connector(\visencoder(\img))$. Higher reconstruction loss indicates greater difficulty in recovering the features that are captured in the original visual embeddings. This patch-level comparison between original and reconstructed embeddings enables us to precisely quantify and locate the visual information loss at a more fine-grained level.
    
    \begin{figure*}[t]
        \centering
        \includegraphics[width=0.9\linewidth]{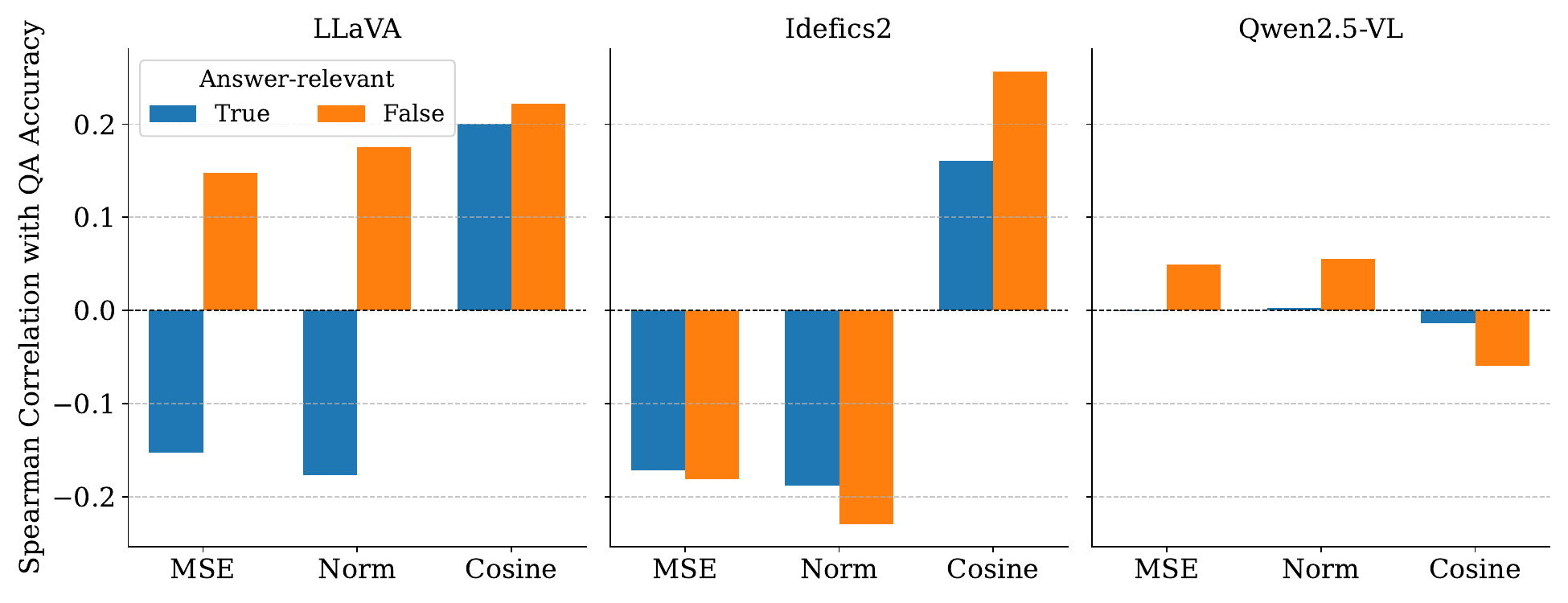}
        \caption{Correlation between reconstruction loss and question-answering accuracy on the VizWiz grounding VQA task. For LLaVA and Idefics2, all correlations have a $p$-value < \num{5e-5}, indicating statistically significant relationships, whereas no clear correlation is observed for Qwen2.5-VL. The reconstruction loss occurs in both answer-relevant and irrelevant patches. Loss in relevant patches negatively affects performance of LLaVA and Idefics2. ``Norm'' represents differences between the $L^2$ norm of the embeddings.}
        \label{fig:neighborhood-overlap-res}
    \end{figure*}
    
    \begin{table}[t!]
    \centering
    \small
    \renewcommand{\arraystretch}{1.1}
    \resizebox{\linewidth}{!}{%
    \begin{tabular}{lrr}
    \toprule
    \textbf{Model} 
    & \textbf{COCO} 
    & \textbf{Flickr30k} \\
    \midrule
    \multicolumn{3}{l}{\textbf{Reconstruction loss (avg / std)
    }} \\
    
    LLaVA      & $0.087\ /\ 0.016$ & $0.097\ /\ 0.019$ \\
    Idefics2    & $0.796\ /\ 0.082$ & $0.854\ /\ 0.074$ \\
    Qwen-2.5-VL & $1.069\ /\ 0.117$  & $1.069\ /\ 0.115$ \\
    \midrule
    \multicolumn{3}{l}{\textbf{Overall CIDEr Scores}} \\
    LLaVA       & $81.28$  & $56.79$ \\
    Idefics2    & $53.64$  & $39.22$ \\
    Qwen-2.5-VL & $13.04$  & $12.85$ \\
    \bottomrule
    \end{tabular}%
    }
    \caption{Reconstruction loss on COCO and Flickr30k test sets. Top: reconstruction loss averaged over all samples, where LLaVA achieves lowest reconstruction error. Bottom: CIDEr scores of zero-shot captioning.\footnotemark For both datasets, we observe better overall captioning performance with lower average reconstruction loss.} 
    \label{tab:captioning}
\end{table}
\footnotetext{We notice Qwen-2.5-VL is particularly sensitive to the task prompt; we use the prompt in the official repo:~\url{https://huggingface.co/Qwen/Qwen2.5-VL-7B-Instruct}.}

\begin{table}[ht]
        \centering
        \small
        \renewcommand{\arraystretch}{1.1}
        \resizebox{\linewidth}{!}{%
        \begin{tabular}{lrr}
        \toprule
        \textbf{Model} 
        & \textbf{COCO} 
        & \textbf{Flickr30k} \\
        \midrule
        \multicolumn{3}{l}{\textbf{CIDEr Scores for High Loss / Low Loss samples}} \\
        LLaVA       & $73.98\ /\ 86.96$  & $51.79\ /\ 61.74$ \\
        Idefics2    & $40.84\ /\ 66.13$  & $29.24\ /\ 53.22$ \\
        Qwen-2.5-VL & $12.45\ /\ 13.56$  & $13.15\ /\ 12.35$ \\
        \midrule
        \multicolumn{3}{l}
        {\textbf{Spearman Correlation} ($\rho$ / $p$)} \\
        LLaVA       & $-0.077\ /\ 0.000$ & $-0.096\ /\ 0.000$ \\
        Idefics2    & $-0.214\ /\ 0.000$ & $-0.226\ /\ 0.000$ \\
        Qwen-2.5-VL & $0.001\ /\ 0.975$  & $0.027\ /\ 0.403$ \\
        \bottomrule
        \end{tabular}%
        }
        \caption{
        Top: The comparison of CIDEr scores for top 25\% highest and 25\% lowest reconstruction loss samples, reported as "High Loss / Low Loss"
        Bottom: Spearman correlations ($\rho$) of per-sample reconstruction loss and captioning CIDEr scores.}
        \label{tab:cap_correlation}
    \end{table}

    \subsection{Reconstruction Loss Impacts Captioning}
    \label{subsec:reconstruction-loss}

        Our embedding reconstruction evaluation follows two steps: 1) we train a reconstruction model for each VLM using paired pre- and post-projection embeddings from images in the COCO 2017 train set (as described in Section~\ref{para:recon-train}); 2) we apply these reconstruction models to predict the original image representations from their projected counterparts.

    For image captioning, we measure the reconstruction loss for images in the Flickr30k validation set and COCO Karpathy test split. We use CIDEr score \citep{vedantam2015ciderconsensusbasedimagedescription} to evaluate the quality of the generated captions.  Table~\ref{tab:captioning} summarizes the overall average reconstruction loss of the three models on the captioning test datasets. For both datasets, we observe lower average reconstruction loss yields better captioning performance. We also investigate how reconstruction loss impacts captioning for each individual image by calculating the correlation between per-sample CIDEr score and reconstruction loss per-image. In Table~\ref{tab:cap_correlation}, the spearman correlation indicates higher reconstruction loss for a given image corresponds to worse captioning for Idefics2 and LLaVA, indicating by the negative correlation with $p$ values smaller than $\num{1e-5}$. Please see more visualization in Figure~\ref{fig:cap-loss}. For Qwen-VL, we did not observe obvious correlation for individual images. The large gap of CIDEr scores between the highest and lowest reconstruction loss samples for LLaVA and Idefics2 suggests substantial impact on downstream tasks.

    \subsection{Loss at Patch-level Visual Features Explains Question Answering Behaviors}
    \label{subsec:res-grounding}
        
        To distinguish whether the reconstruction loss stems from selective feature preservation or actual information loss, we visualize the patch-level loss for images in the VizWiz grounding VQA validation dataset. This dataset is particularly suitable for our analysis as it provides answer grounding—binary masks indicating image regions relevant to each question. By examining the relationship between the reconstruction loss for the answer-relevant image patches and question-answering accuracy, we can assess whether the projection preserves task-relevant visual information.

        We report the Spearman correlation between the reconstruction loss and the question answering accuracy in Figure~\ref{fig:neighborhood-overlap-res}. For LLaVA, we observe a negative correlation between prediction accuracy and reconstruction loss in answer-relevant patches, while a positive correlation is found in irrelevant patches. This indicates that information loss in answer-relevant patches negatively impacts model performance, whereas loss in irrelevant patches has a less significant effect. For Idefics2, we can see that information loss in any patches would hurt question answering accuracy. We do not observe significant correlation for Qwen-2.5-VL, which is consistent with our findings in the captioning tasks.
        
        As shown in Figure~\ref{fig:patch_differences}, identifying distorted features allows us to pinpoint visual information that becomes inaccessible or less reliable for the language model. For instance, reconstruction loss in the patches of the fifth number "8" rank among the top ten of all image patches, suggesting that the model may have struggled to answer the question due to lost details necessary for identifying the number. This analysis introduces a new visualization approach to examine \textsc{VLM} limitations, particularly in scenarios requiring reasoning or recognizing fine-grained viusal features. Please see more visualization examples in Appendix~\ref{app:vis}.

\section{Analysis}
\label{sec:analysis}
    \paragraph*{Procrustes analysis}
    \label{subsec:procrustes}
    We also attempt to find the optimal geometrical transformation from the post-projection embedding space to the pre-projection one through Procrustes analysis \cite{gower1975generalized} -- a method often used for supervised alignment of embeddings~\cite{artetxe2018generalizing}. The alignment error reflects the degree of structural similarity of the two embedding spaces. We use mean-pooled image embeddings from LLaVA, Idefics2, and Qwen2.5-VL. As the pre- and post-projection embeddings have different embedding dimensions and sequence lengths, our analysis follows three steps to complete the embedding alignment. We first take the mean-pooled image representation by averaging over the sequence length, producing fixed-size vectors of size $\visembdim$ and $\embdim$. We then use PCA~\cite{hotelling1933analysis} on the mean-pooled post-projection embeddings to project them to the same dimension of the mean-pooled pre-projection embeddings.
    
    Orthogonal transformation matrix $\mathbf{R}$ was derived through singular value decomposition of the cross-covariance matrix $\bar{X}^\top \bar{T}$, where $\bar{X} \in \mathbb{R}^{\visembdim}$ represents mean-pooled pre-projection embeddings and $\bar{T} \in \mathbb{R}^{\visembdim}$ the PCA-transformed post-projection embeddings. Then the orthogonal transformation matrix is learned to best align these two sets of embeddings by minimizing the Euclidean distance. The reconstruction error are reported in Table~\ref{tab:procrustes_errors}. Figure~\ref{fig:llava-procrustes} visualizes the alignment of LLaVA embeddings through procrustes analysis.
    
    \begin{table}[t]
        \centering
        \resizebox{0.8\columnwidth}{!}{%
        \begin{tabular}{lrrrr}
            \toprule
            \textbf{Model} & \textbf{Mean} & \textbf{Std} & \textbf{Min} & \textbf{Max} \\
            \midrule
            LLaVA     & 16.62  & 3.16  & 8.76  & 23.65 \\
            Idefics2  & 4.93   & 0.08  & 4.78  & 5.70 \\
            Qwen2.5-VL & 4.41 & 0.09 & 4.24 & 5.05 \\
            \bottomrule
        \end{tabular}%
        }
        \caption{Procrustes analysis results. We report the alignment error on SeedBench image representations before and after connector projection.}
        \label{tab:procrustes_errors}
    \end{table}
    
    Our analysis reveals fundamental limitations in linear alignment of the image embeddings. The high alignment errors of 16.62 for LLaVA and 4.41 for Qwen2.5-VL indicate the inherent difficulty of preserving geometric relationships through rigid transformations. While serving as a critical baseline for structural fidelity assessment, this constrained linear approach explains why our proposed non-linear embedding reconstruction approach achieves significantly lower errors.

    \paragraph*{Ablation on Reconstruction Model Size and Structure}
    \label{para:ablation}
    We train three reconstruction models of different sizes for LLaVA: a 27M three-layer MLP, a 39M five-layer MLP, and a 40M Transformer. In Table~\ref{tab:size_ablation}, we observe that the 27M model is sufficient for reconstructing LLaVA visual embeddings, and a larger model does not yield better validation loss.

    \section{Related Work}
    \label{sec:related}
    A series of analyses has been conducted to investigate the modality gap and representation limitations of contrastive-based VLMs \citep{schrodi2024two, ModalityGap, mmvp}. These studies reveal that the representational shortcomings in CLIP embeddings subsequently impact the visual perception capabilities of VLMs relying on such vision encoders. For connector-based VLMs, \citet{VLMClassifier} demonstrates that the latent space sufficiently retains the information necessary for classification through probing across different layers, and \citet{lin-etal-2024-preserve} demonstrates the impact of different connectors on VLMs' downstream performance. However, there remains a significant gap in understanding whether fine-grained visual information, crucial for tasks such as visual grounding~\cite{VisualGenomeConnecting2017a} and question answering~\cite{chen2022grounding}, is lost in the process. In this paper, we focus on the connector-based models to understand the information transformation. To the best of our knowledge, our paper is the first to directly quantify information loss of the connectors from the representation perspective, offering deeper insights into where and what specific information is lost from the visual features.

    \begin{table}[t!]
        \centering
        \resizebox{\columnwidth}{!}{%
        \begin{tabular}{llrrr}
        \toprule
        \textbf{Model} & \textbf{Size} & \textbf{VizWiz} & \textbf{SeedBench} & \textbf{FoodieQA} \\
        \midrule
        \multirow{2}{*}{MLP} & \multirow{2}{*}{27M} & Avg 0.050 & 0.056 & 0.051 \\
                             &                      & Std 0.013 & 0.011 & 0.007 \\
        \midrule
        \multirow{2}{*}{MLP} & \multirow{2}{*}{39M} & Avg 0.064 & 0.070 & 0.065 \\
                             &                      & Std 0.015 & 0.013 & 0.008 \\
        \midrule
        \multirow{2}{*}{Transformer} & \multirow{2}{*}{40M} & Avg 0.237 & 0.231 & 0.228 \\
                                     &                      & Std 0.019 & 0.025 & 0.014 \\
        \bottomrule
        \end{tabular}%
        }
        \caption{Reconstruction loss with different architectures across VizWiz, SeedBench, and FoodieQA datasets. Reported values include average loss (Avg) and standard deviation (Std).}
        \label{tab:size_ablation}
    \end{table}

\section{Conclusion and Future Work}
    Our study provides a systematic evaluation of how connectors in existing VLMs transform information and reshape the representation space when projecting visual embeddings into the language embedding space. Through neighborhood overlap ratios and embedding reconstruction, we establish a quantitative framework that captures two critical aspects of the information loss: 1) structural shift of global semantic relationships shown by the 40-60\% divergence in nearest-neighbor rankings
    and 2) patch-level reconstruction loss negatively impacts captioning performance and explains model failures in fine-grained visually grounded question answering. The patch-level reconstruction also enables visualization of local information loss, offering interpretable explanations for model behaviors.
    
    Our findings suggest two key properties of an effective connector: preserving or improving semantic representation of images and preserving visual information most relevant to the text context. These findings could guide further improvements in VLM connectors. For example, the reconstruction loss at the embedding level could potentially be incorporated during model pretraining as regularization.  Future work could also explore designing dynamic projection layers or better visual feature selection mechanisms for modality fusion.

\section*{Ethics Statement}
    We foresee no ethical concerns with our research project. In particular, ours is merely a scientific study of VLMs and provides no artifacts that can be used in a real-world scenario.

\section*{Limitations}
    In this study, we evaluate the information loss introduced by connectors in VLMs. However, several limitations should be noted. First, due to variations in model architectures and pretraining strategies, our findings may be specific to the connector-based VLMs analyzed and may not generalize to architectures that employ cross-attention for modality fusion. Second, our experiments focus on connectors in VLMs within the 7B–8B parameter range. Expanding the analysis to models of different sizes could provide deeper insights into the relationship between model scale and information loss. Third, our pixel-level reconstruction experiments (Appendix~\ref{app:image_reconstruct}) yielded inconclusive results in quantifying information loss, possibly due to limitations in our chosen image generation model and training dataset size. Additionally, while we empirically validate our $k$-NN overlap ratio and embedding reconstruction metrics, a formal theoretical characterization would further strengthen their reliability. Finally, our reconstruction experiments cannot conclusively determine whether the observed information loss stems from the connector layer itself or from potential learning limitations of the trained reconstruction network.

\section*{Acknowledgments}
 Wenyan Li is supported by the Lundbeck Foundation (BrainDrugs grant: R279-2018-1145). We sincerely appreciate the help and feedback provided by Ryan Cotterell, Vésteinn Snæbjarnarson, and Clemente Pasti from the Rycolab at ETH Zürich, especially regarding the formal mathematical formulations.

\bibliography{anthology,custom}
\bibliographystyle{acl_natbib}

\appendix
\onecolumn
\appendix

\section{Connectors in Autoregressive Vision-Language Models}
\label{sec:connectors}

\paragraph{Idefics2}
Idefics2 leverages a perceiver resampler~\cite{jaegle2021perceiver} as the connector. The perceiver resampler forms an attention bottleneck that encourages the latent representations to attend to the most relevant inputs in a high-dimensional input array through iterative cross-attention layers. In other words, the cross-attention module projects the high-dimensional inputs into a fixed-dimensional learned representation. Please refer to \citet{laurençon2024matters} for more details.

\paragraph{LLaVA}
LLaVA~\cite{liu2023llava} uses a two layer MLP to project the image embeddings to the language model's embeddings space. The MLP projector preserves the image feature length -- number of patches extracted by the image encoder.

\paragraph{Qwen2.5-VL}
Qwen2.5-VL~\cite{bai2025qwen25vltechnicalreport} uses a patch merger (two-layer MLP) to reduces the length of the input image features. The image representations of the  neighboring four patches in the image are first merged, and then passed through a two-layer MLP to project the image representation to the LM embedding dimension.

\section{Ablation on Index Method for $k$-NN Overlap Ratio}
We evaluated $k$-NN overlap ratio using three different embedding types as search indices: original embeddings, mean-pooled image embeddings, and normalized embeddings (Table~\ref{tab:index}). Since the performance differences were minimal, we selected mean-pooled embeddings for both pre- and post-projection image representations in calculating $k$-NN overlap ratios.

\begin{table*}[h]
    \centering
    \small
    \begin{tabular}{l *{6}{S[table-format=1.3]}}
    \toprule
    \multirow{3}{*}{\textbf{Overlap Ratio}} & \multicolumn{6}{c}{\textbf{Index Type}} \\
    \cmidrule(lr){2-7}
    & \multicolumn{2}{c}{IndexFlatL2} & \multicolumn{2}{c}{\makecell{IndexFlatL2\\(mean pooling)}} & \multicolumn{2}{c}{\makecell{IndexFlatIP\\(normalized vectors)}} \\
    \cmidrule(lr){2-3} \cmidrule(lr){4-5} \cmidrule(lr){6-7}
    & {\textbf{mean}} & {\textbf{std}} & {\textbf{mean}} & {\textbf{std}} & {\textbf{mean}} & {\textbf{std}} \\
    \midrule
    top100  & 0.466 & 0.122 & 0.563 & 0.107 & 0.504 & 0.129 \\
    top50  & 0.488 & 0.128 & 0.556 & 0.120 & 0.425 & 0.142 \\
    top10  & 0.490 & 0.149 & 0.551 & 0.160 & 0.377 & 0.161 \\
    \midrule
    \multicolumn{7}{l}{\textbf{Vector Size}} \\
    Before projection & \multicolumn{2}{c}{576$\times$1024} & \multicolumn{2}{c}{1$\times$1024} & \multicolumn{2}{c}{576$\times$1024} \\
    After projection & \multicolumn{2}{c}{576$\times$4096} & \multicolumn{2}{c}{1$\times$4096} & \multicolumn{2}{c}{576$\times$4096} \\
    \bottomrule
    \end{tabular}
    \caption{Ablation on KNN results when using original embeddings, mean pooled image embeddings, and normalized embeddings. We chose to use the mean-pooled embeddings for efficiency due to large embeddings size.}
    \label{tab:index}
\end{table*}

\section{Additional Evaluation Results}
\subsection{CUB image retrieval performance}
In Table~\ref{tab:cub_res}, we show the image retrieval performance on CUB test set using $L^2$ and inner product for similarity measure. The performance are consistent regardless of the index method used.

\subsection{Reconstruction loss on VQA datasets}
    For visual question answering tasks, we measure the reconstruction loss for images in the validation set of VizWiz grounding VQA, Seed-Bench, and FoodieQA.
    Table~\ref{tab:mse_loss} presents overall reconstruction loss. Among all tested models, LLaVA's projected embeddings maintain the highest reconstruction fidelity. The overall reconstruction loss reflects the overall difficulty of recovering information encoded in the visual representations. 
  
\begin{table}[h]
    \begin{minipage}[t]{0.48\textwidth}
        \vspace{0pt}  %
        \centering
        \adjustbox{max width=\textwidth}{
        \begin{tabular}{l rrrr}
        \toprule
        \textbf{Model} & \multicolumn{2}{c}{\textbf{L2}} & \multicolumn{2}{c}{\textbf{IP}} \\
        \cmidrule(lr){2-3} \cmidrule(lr){4-5}
        & \textbf{R@1} & \textbf{R@5} & \textbf{R@1} & \textbf{R@5} \\
        \midrule
        \multicolumn{5}{l}{\textit{Pre-projection}} \\
        LLaVA & 8.34 & 21.82 & 9.46 & 24.78 \\
        Idefics2 & \textbf{13.10} & \textbf{30.81} & \textbf{13.38} & \textbf{30.98} \\
        Qwen-2.5-VL & 4.23 & 11.74 & 6.83 & 24.23 \\
        \midrule
        \multicolumn{5}{l}{\textit{Post-projection}} \\
        LLaVA & 6.16 ↓ & 17.22 ↓ & 5.54 ↓ & 20.49 ↓ \\
        Idefics2 & \textbf{10.87} ↓ & 25.28 ↓ & \textbf{10.99} ↓ & 25.15 ↓ \\
        Qwen-2.5-VL & 10.65 ↑ & \textbf{26.44} ↑ & 8.26 ↑ & \textbf{26.70} ↑ \\
        \bottomrule
        \end{tabular}
        }
        \caption{Zero-shot retrieval performance on CUB test set using $L^2$ distance and inner product for similarity measure. R@$k$ denotes Recall at rank $k$. Arrows indicate performance change direction after projection.}
        \label{tab:cub_res}
    \end{minipage}%
    \hfill%
    \begin{minipage}[t]{0.48\textwidth}
        \vspace{0pt}  %
        \centering
        \adjustbox{max width=\textwidth}{
        \begin{tabular}{l c c c c}
        \toprule
        \textbf{Dataset} & \textbf{MSE} & \textbf{LLaVA} & \textbf{Idefics2} & \textbf{Qwen2.5-VL} \\
        \midrule
        \multirow{2}{*}{VizWiz} & Avg & \textbf{0.115} & 0.907 & 1.069 \\
         & Std & 0.086 & 0.298 & 0.684 \\
        \midrule
        \multirow{2}{*}{SeedBench} & Avg & \textbf{0.106} & 0.872 & 1.069 \\
         & Std & 0.071 & 0.307 & 0.610 \\
        \midrule
        \multirow{2}{*}{FoodieQA} & Avg & \textbf{0.113} & 0.918 & 1.069 \\
         & Std & 0.057 & 0.283 & 0.673 \\
        \bottomrule
        \end{tabular}
        }
        \vspace{0pt}
        \caption{Embedding reconstruction loss of images in the VizWiz, SeedBench, and FoodieQA datasets. We report both average loss (avg) and standard deviation (std). LLaVA's visual embeddings exhibit lowest reconstruction error among all models. The reconstruction performance is consistent to what we have observed for the images in COCO and Flickr30k.}
        \label{tab:mse_loss}
    \end{minipage}
\end{table}

\section{Visualization}
\label{app:vis}

\subsection{Visualization for Procrustes Analysis}
\begin{figure}[th]
    \centering
    \vspace{5em}
        \includegraphics[width=0.6\linewidth, trim={15em 2em 5em 5em, clip}]{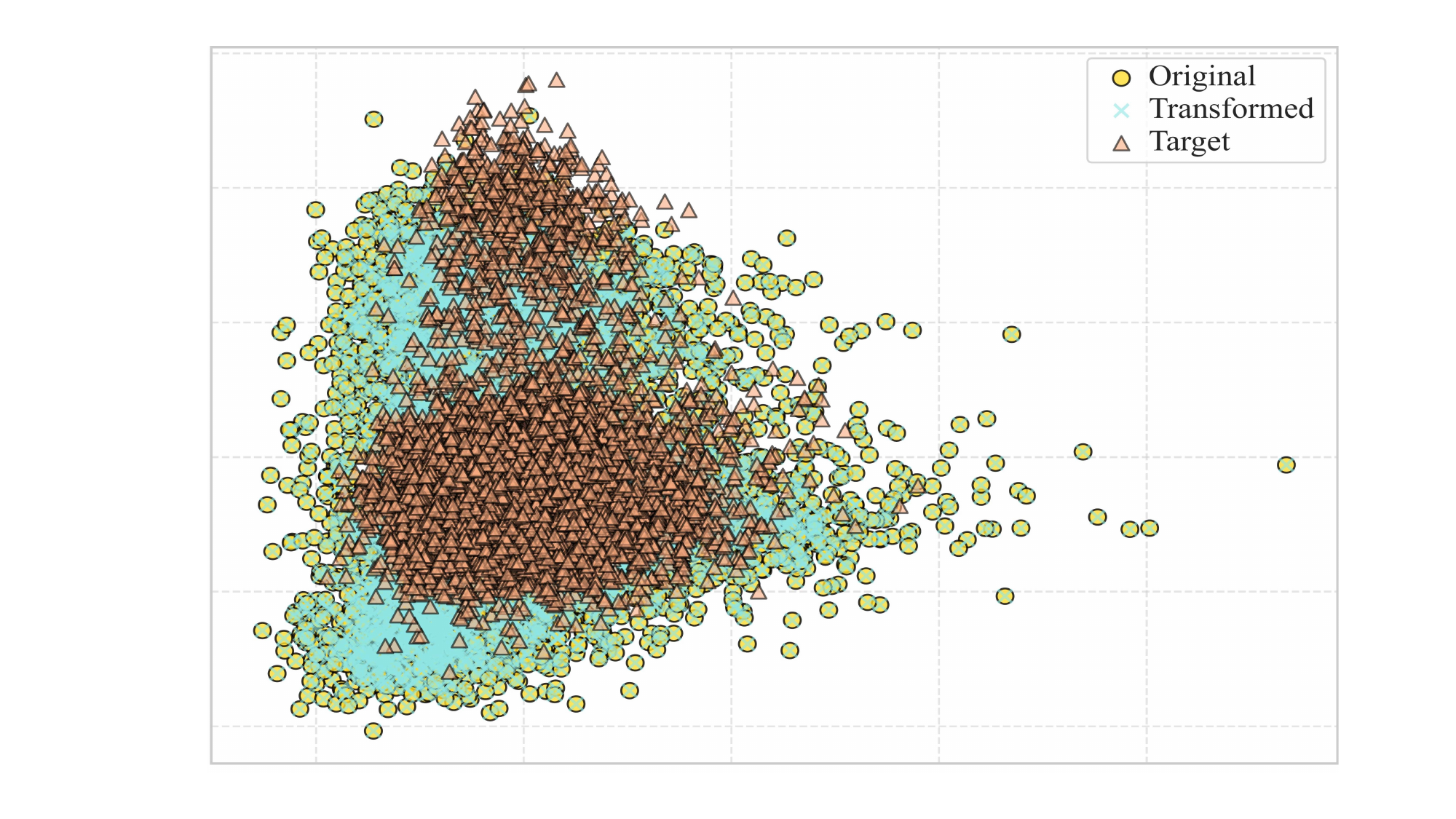} \\
        \caption{Alignment visualization for LLaVA pre- and post-projection embeddings through PCA.}
        \label{fig:llava-procrustes}
\end{figure}

In Figure~\ref{fig:llava-procrustes}, we visualize the alignment for LLaVA pre- and post-projection embeddings, as well as the embeddings learned through the linear transformation learned. From the visualization we can observe that the linear transformation is not able to align the pre- and post-projection embeddings well.

\subsection{Patch-level Loss Visualization for Vizwiz Grounding VQA}
\begin{figure*}[t]
    \centering
    \begin{subfigure}[b]{0.32\textwidth}
        \centering
        \includegraphics[width=\textwidth, trim={0pt 128pt 240pt 128pt}, clip]{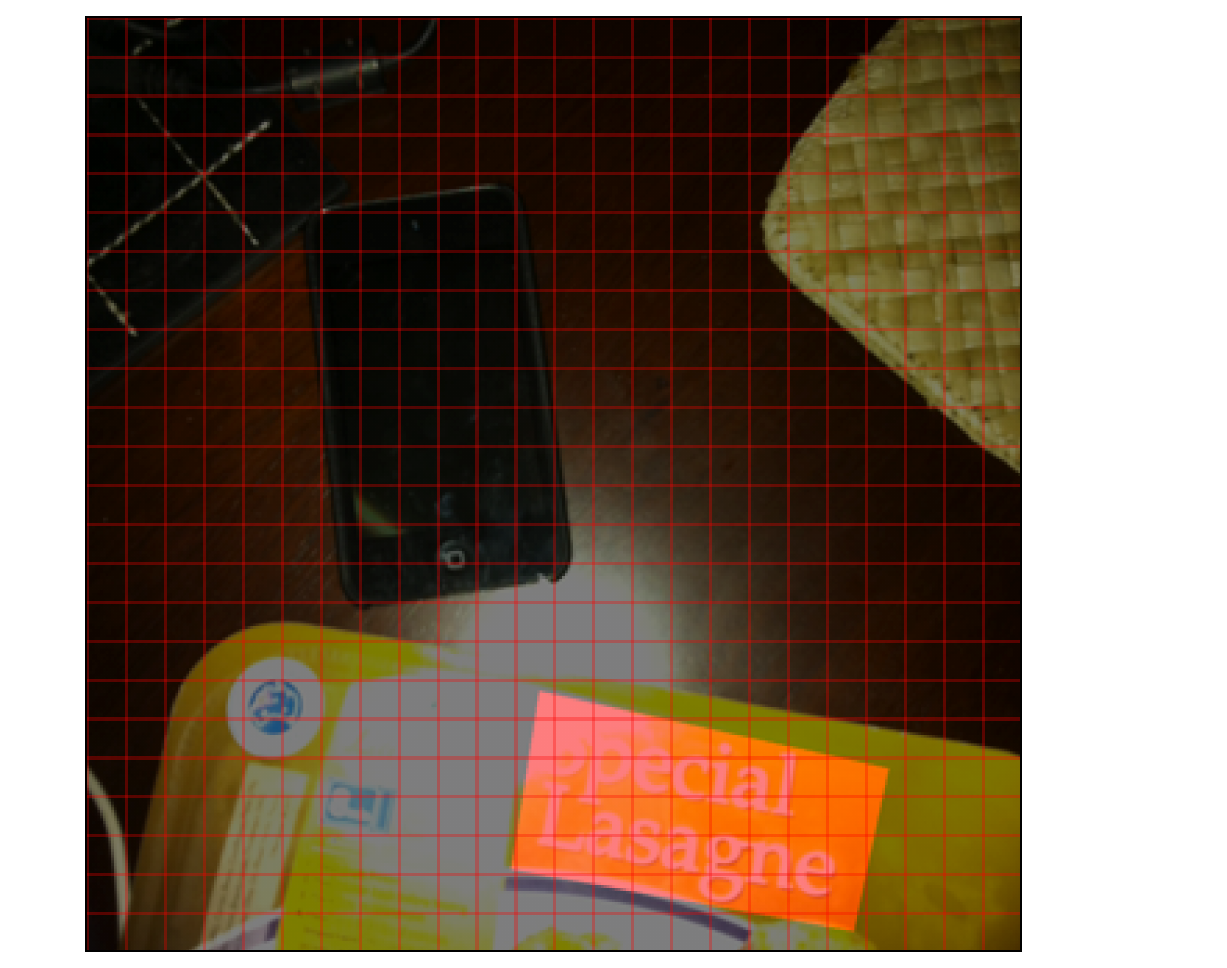}
    \end{subfigure}
    \hfill
    \begin{subfigure}[b]{0.32\textwidth}
        \centering
        \includegraphics[width=\textwidth, trim={40pt 128pt 200pt 128pt}, clip]{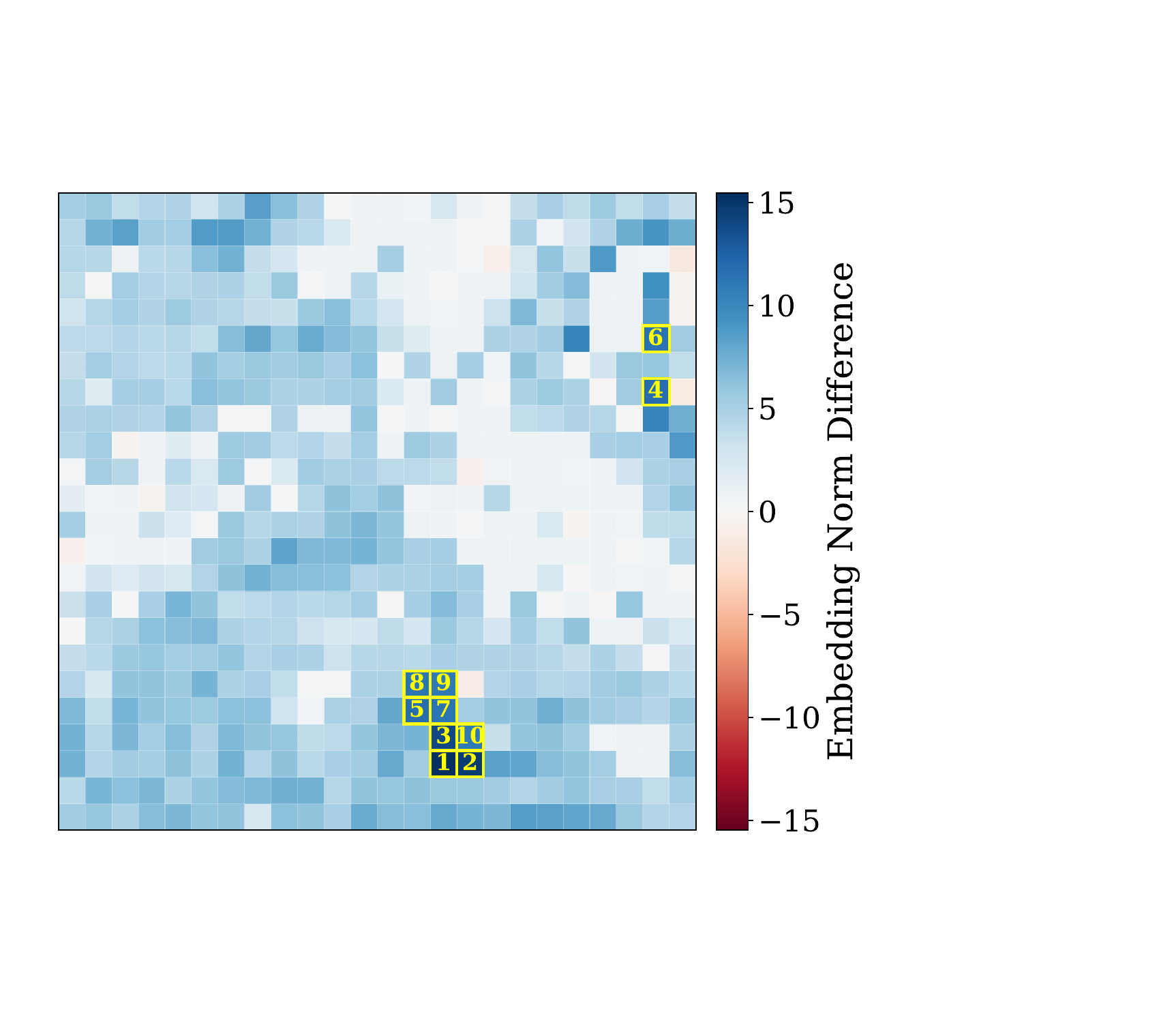}
    \end{subfigure}
    \hfill
    \begin{subfigure}[b]{0.32\textwidth}
        \centering
        \includegraphics[width=\textwidth, trim={40pt 128pt 200pt 128pt}, clip]{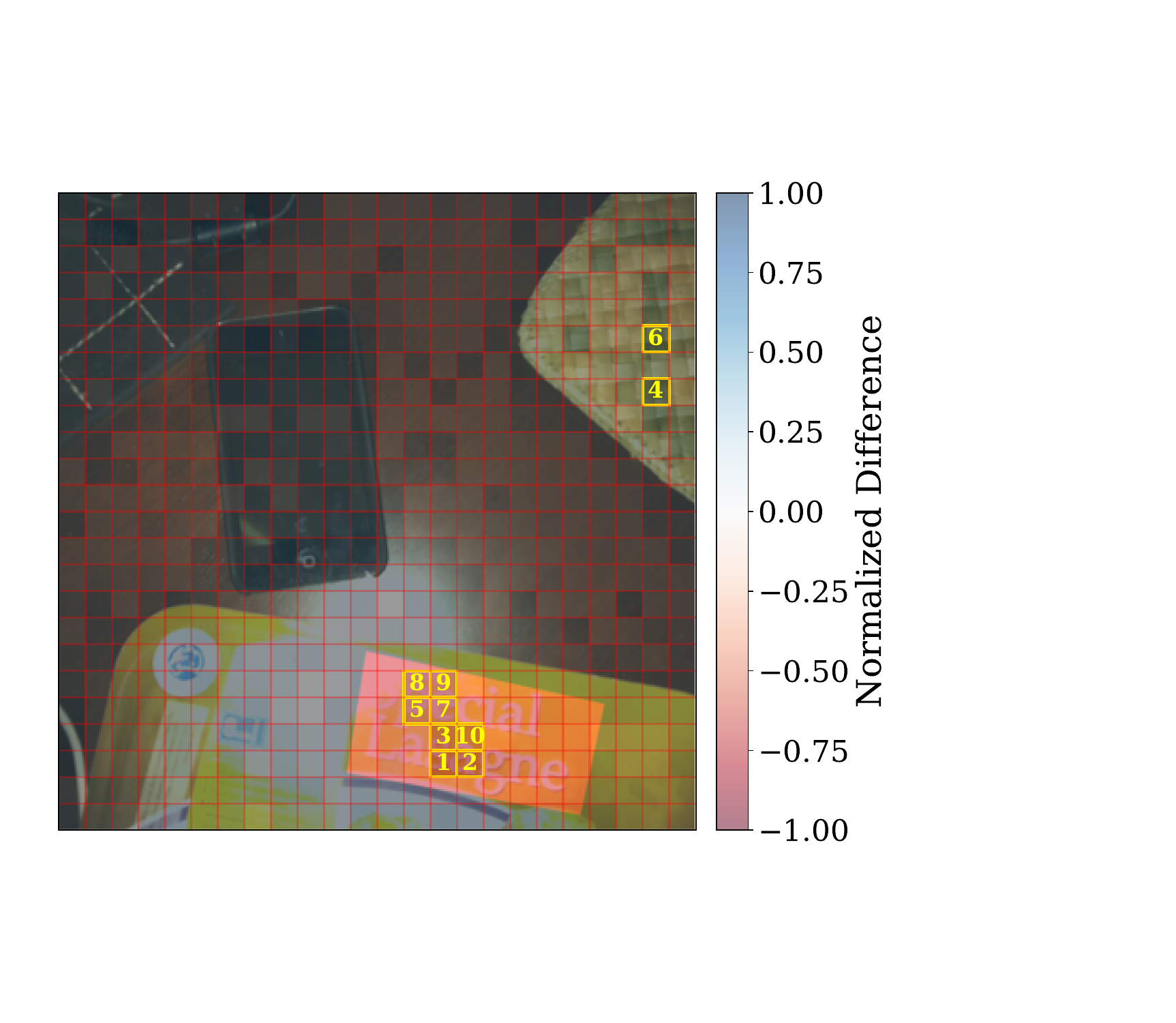}
    \end{subfigure}
    \\[\baselineskip]
    \begin{subfigure}[b]{0.32\textwidth}
        \centering
        \includegraphics[width=\textwidth, trim={0pt 128pt 240pt 128pt}, clip]{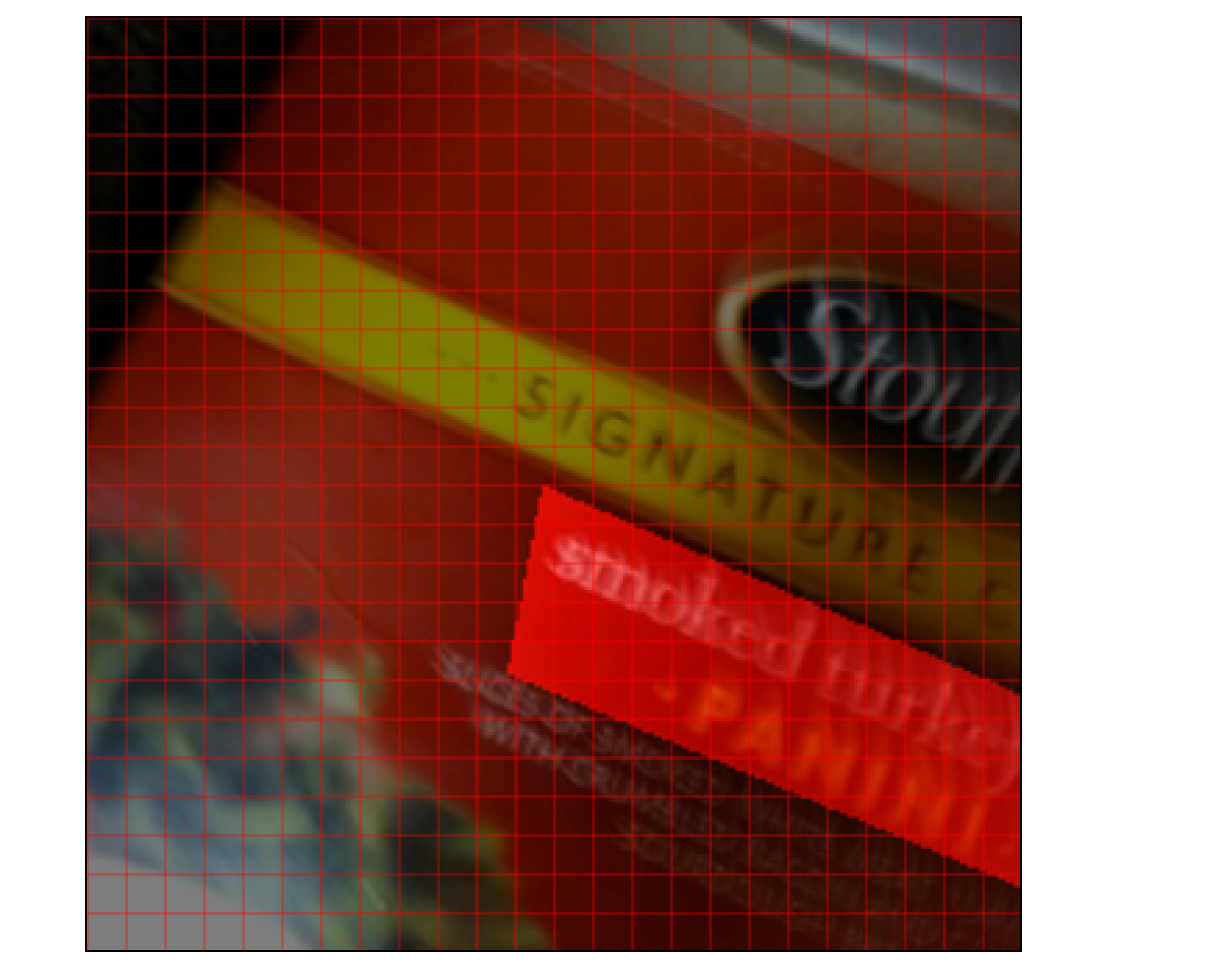}
    \end{subfigure}
    \hfill
    \begin{subfigure}[b]{0.32\textwidth}
        \centering
        \includegraphics[width=\textwidth, trim={40pt 128pt 200pt 128pt}, clip]{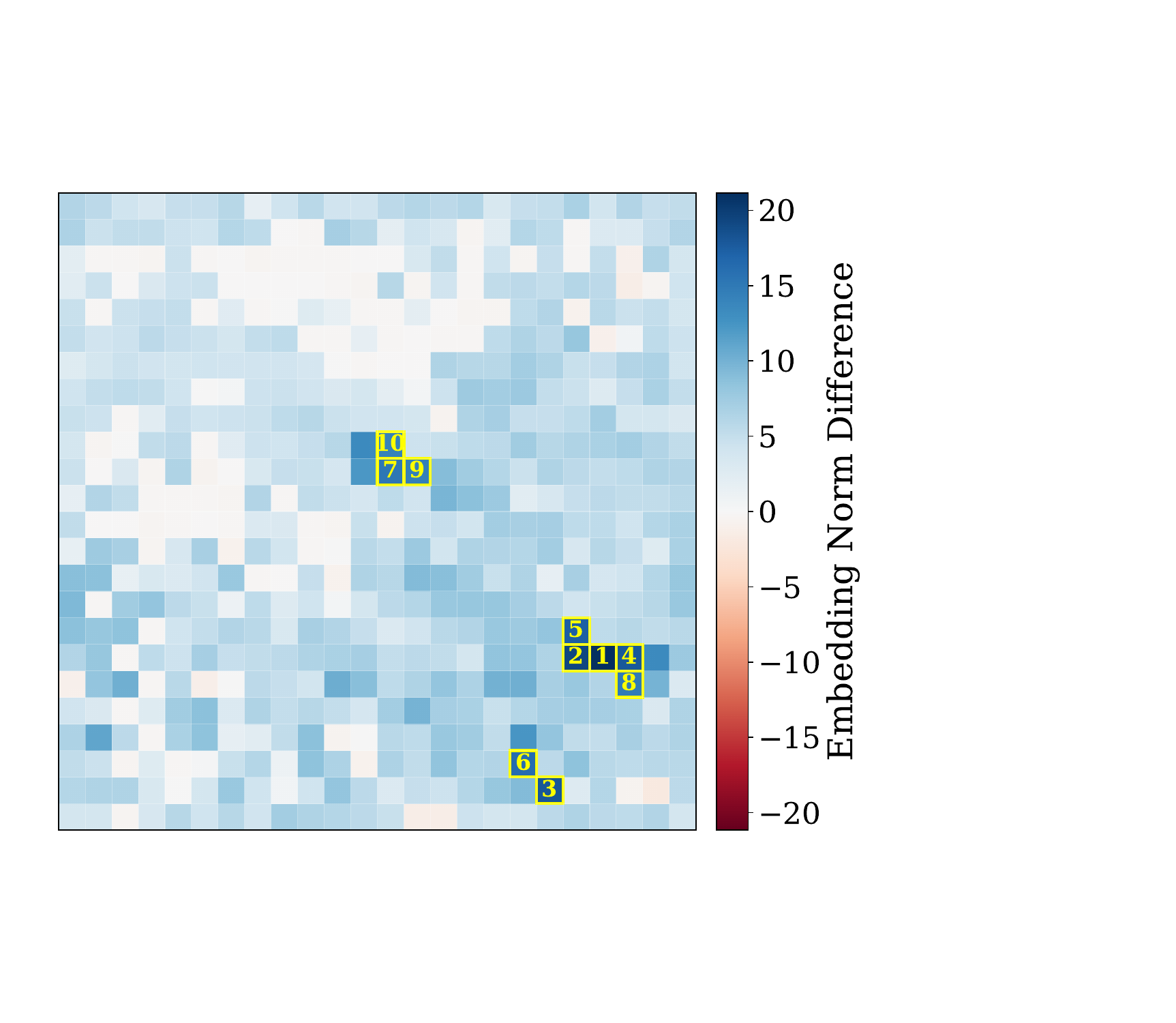}
    \end{subfigure}
    \hfill
    \begin{subfigure}[b]{0.32\textwidth}
        \centering
        \includegraphics[width=\textwidth, trim={40pt 128pt 200pt 128pt}, clip]{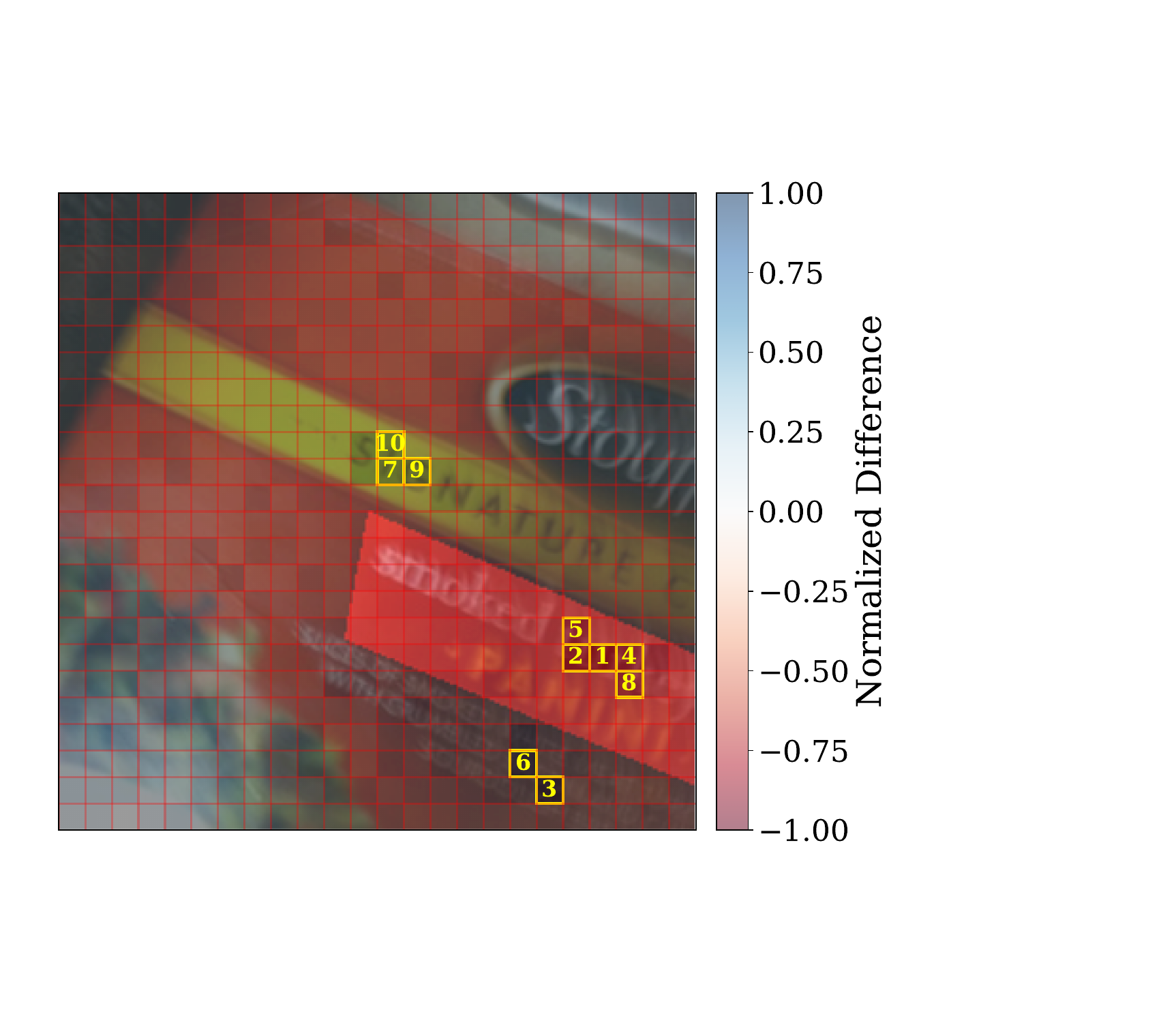}
    \end{subfigure}
    \\[\baselineskip]
    \begin{subfigure}[b]{0.32\textwidth}
        \centering
        \includegraphics[width=\textwidth, trim={0pt 128pt 240pt 128pt}, clip]{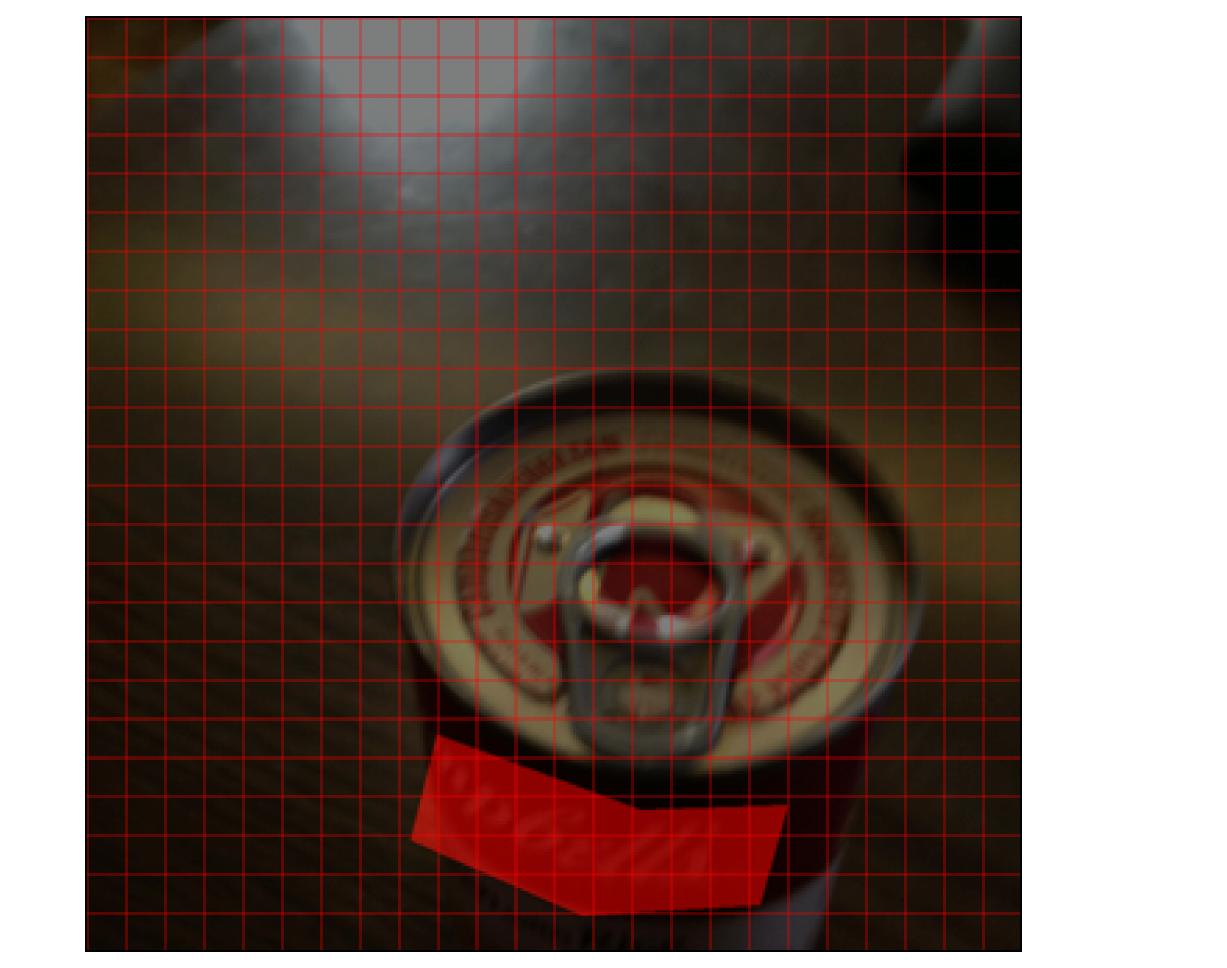}
    \end{subfigure}
    \hfill
    \begin{subfigure}[b]{0.32\textwidth}
        \centering
        \includegraphics[width=\textwidth, trim={40pt 128pt 200pt 128pt}, clip]{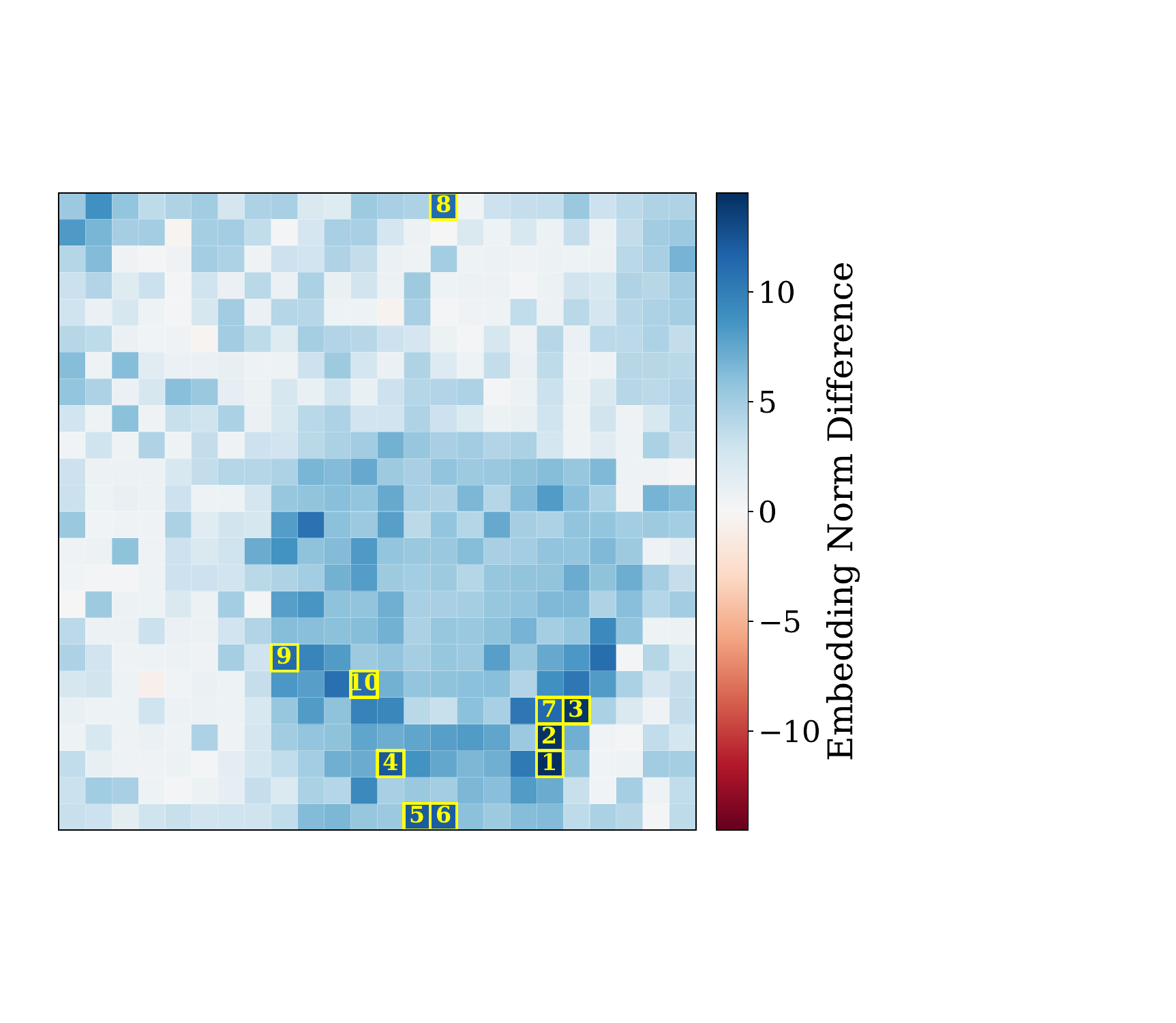}
    \end{subfigure}
    \hfill
    \begin{subfigure}[b]{0.32\textwidth}
        \centering
        \includegraphics[width=\textwidth, trim={40pt 128pt 200pt 128pt}, clip]{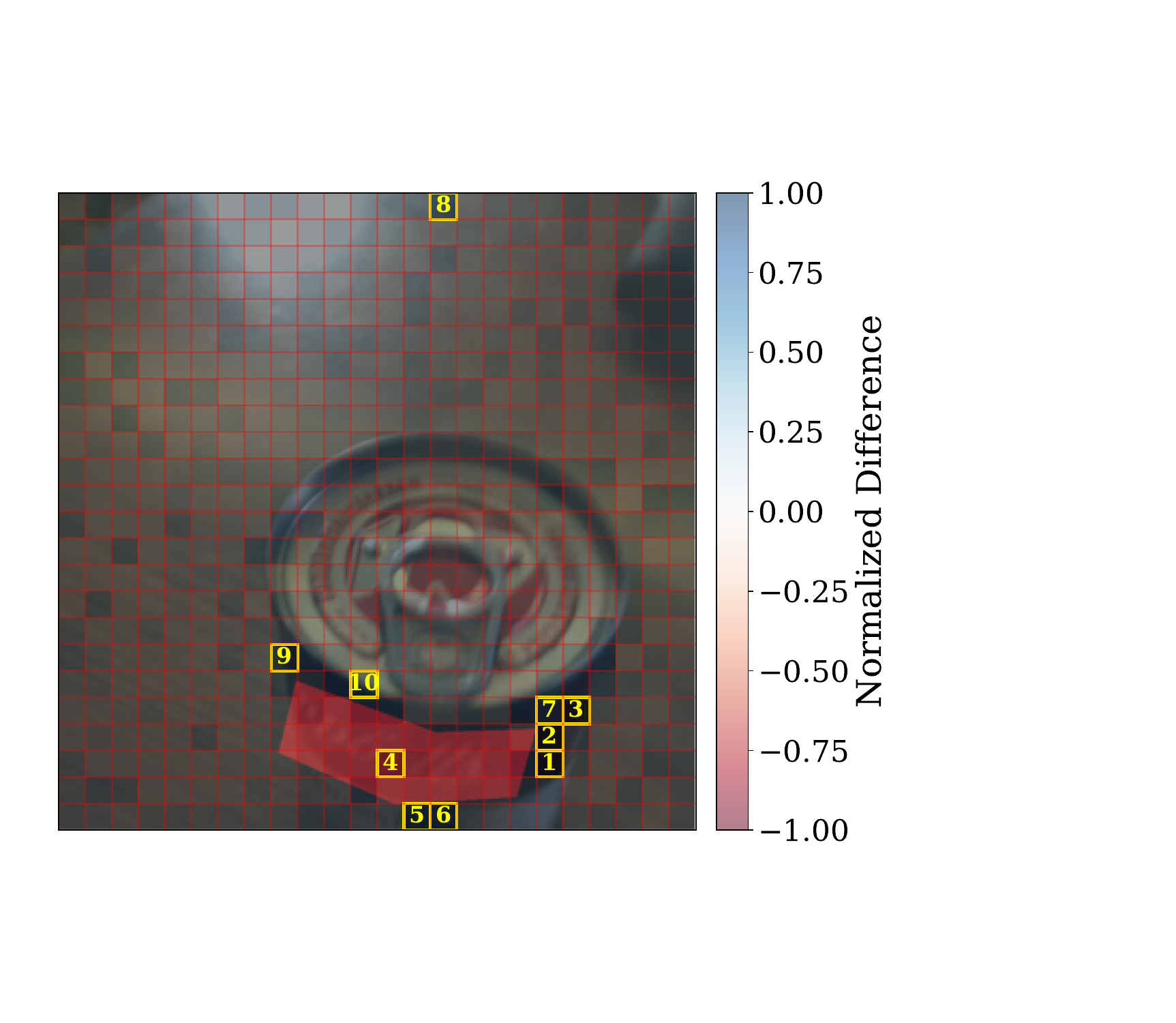}
    \end{subfigure}
    \caption{Additional visualization of high reconstruction loss patches that contributes to model's failure on answering questions that requires recognizing text in the objects. Left: input images with answer-relevant regions in red masks. Middle: signed difference between post-projection embeddings norms and pre-projection embedding norms. Right: normalized norm differences overlay with the input image, with highest loss patches marked in yellow.}
    \label{fig:patch-losses}
\end{figure*}

In Figure~\ref{fig:patch-losses}, we visualize additional examples of high reconstruction loss patches that contributes to model's failure on answering questions that requires recognizing text in the objects.

\begin{figure*}[thb]
    \centering
    \includegraphics[width=\linewidth]{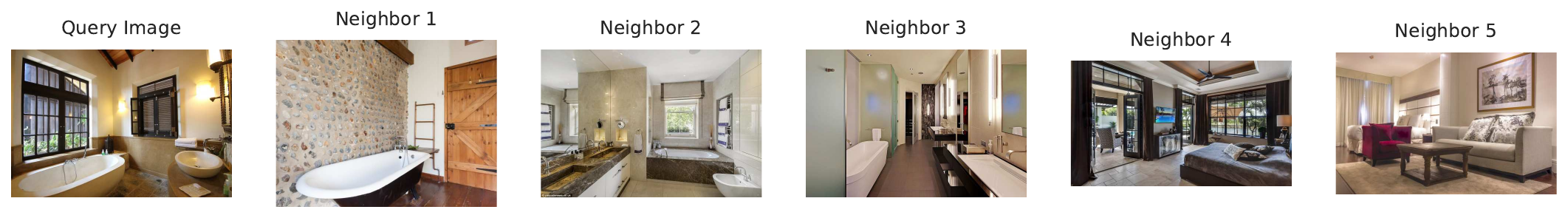}
    \includegraphics[width=\linewidth]{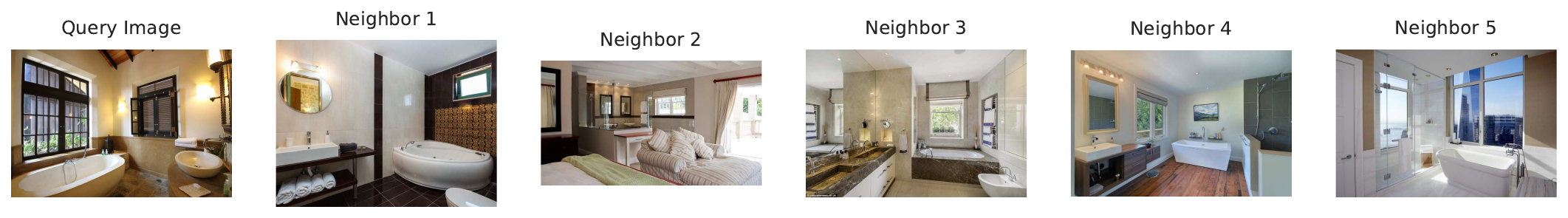}
    \caption{Idefics high $k$NN overlap ratio example, where we can observe the reordering among semantically similar vision embeddings.}
    \includegraphics[width=\linewidth]{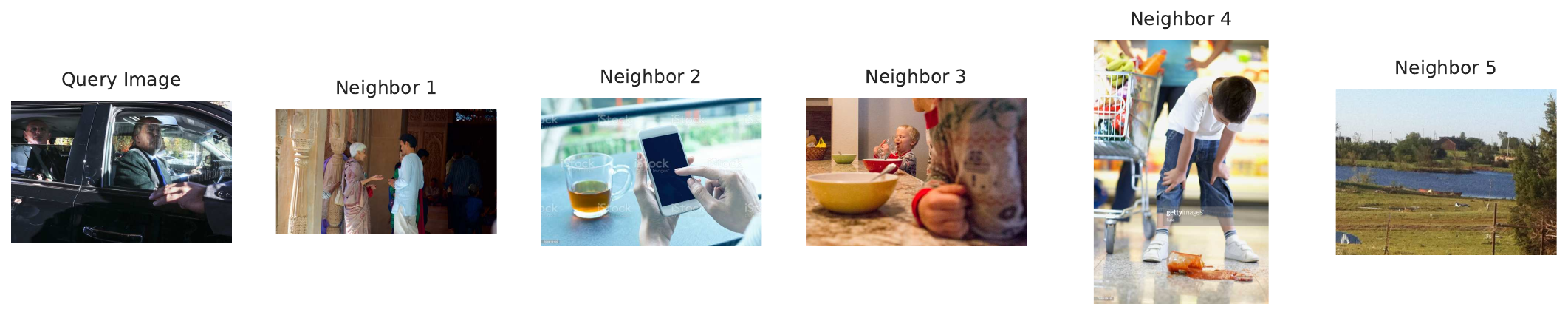}
    \includegraphics[width=\linewidth]{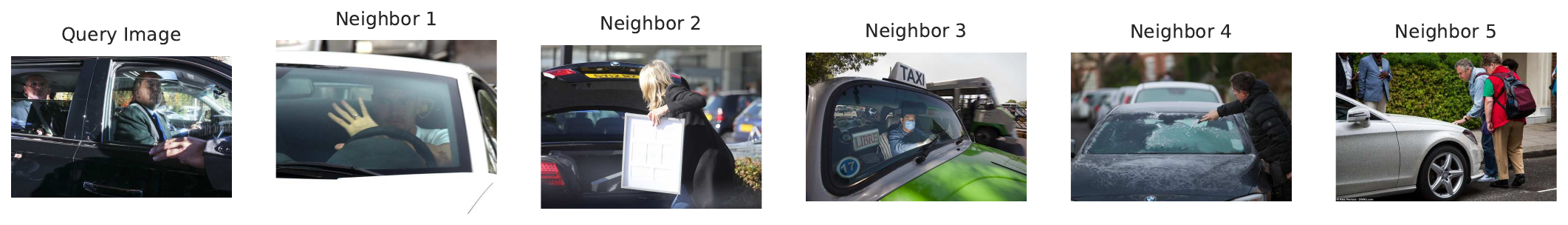}
     \caption{Qwen $k$NN example where the post-projection embeddings are better at retrieving semantically similar images (bottom).}
     \includegraphics[width=\linewidth]{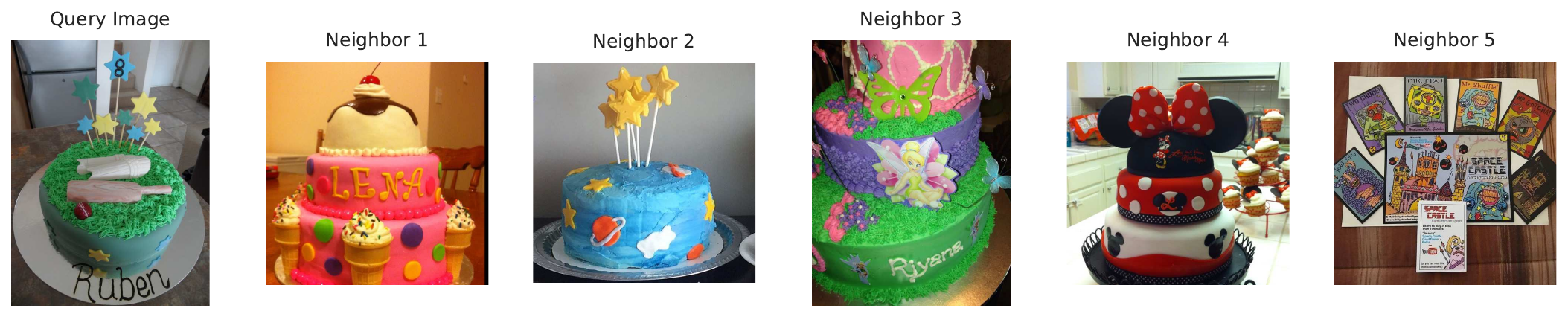}
     \includegraphics[width=\linewidth]{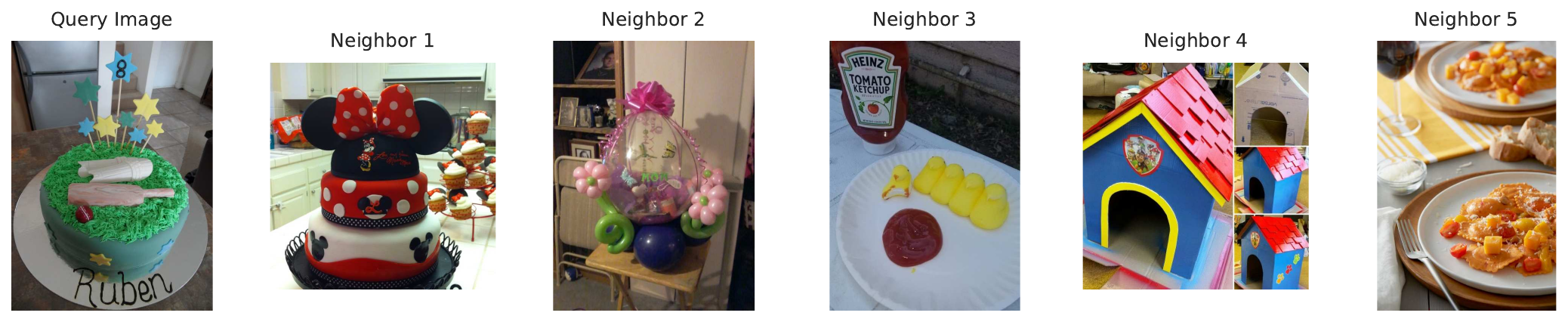}
     \caption{LLaVA low $k$NN overlap ratio example. We can observe the degradation in post-projection embedding.}
    \label{fig:knns}
\end{figure*}

\subsection{Visualization of Neighborhood Reordering}
\begin{figure}[thb]
        \centering
    
        \begin{subfigure}[b]{\columnwidth}
            \centering
            \includegraphics[width=0.8\columnwidth]{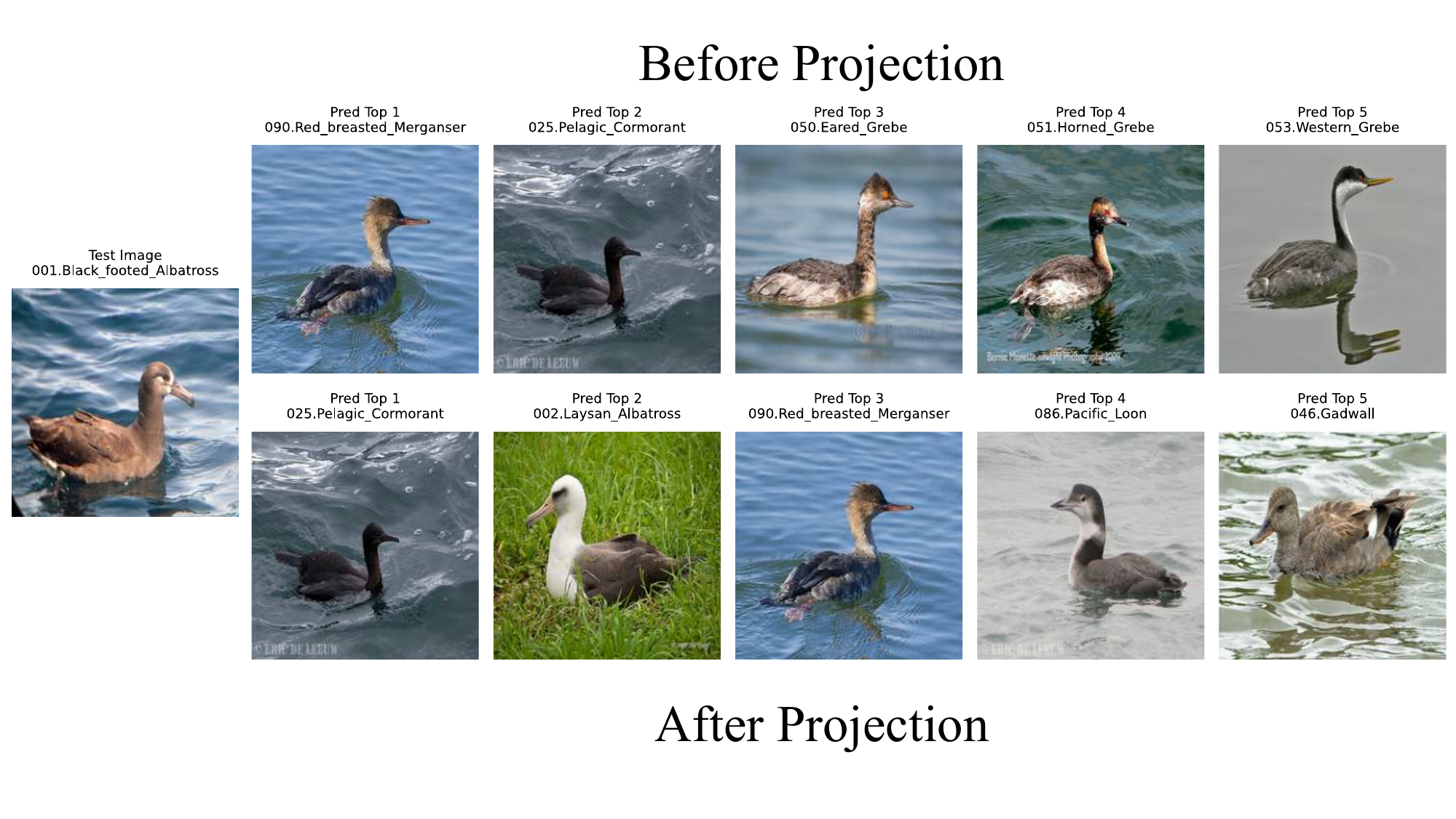}
            \caption{Top five retrieved images of LLaVA image embeddings}
        \end{subfigure}

        \begin{subfigure}[b]{\columnwidth}
            \centering
            \includegraphics[width=0.8\columnwidth]{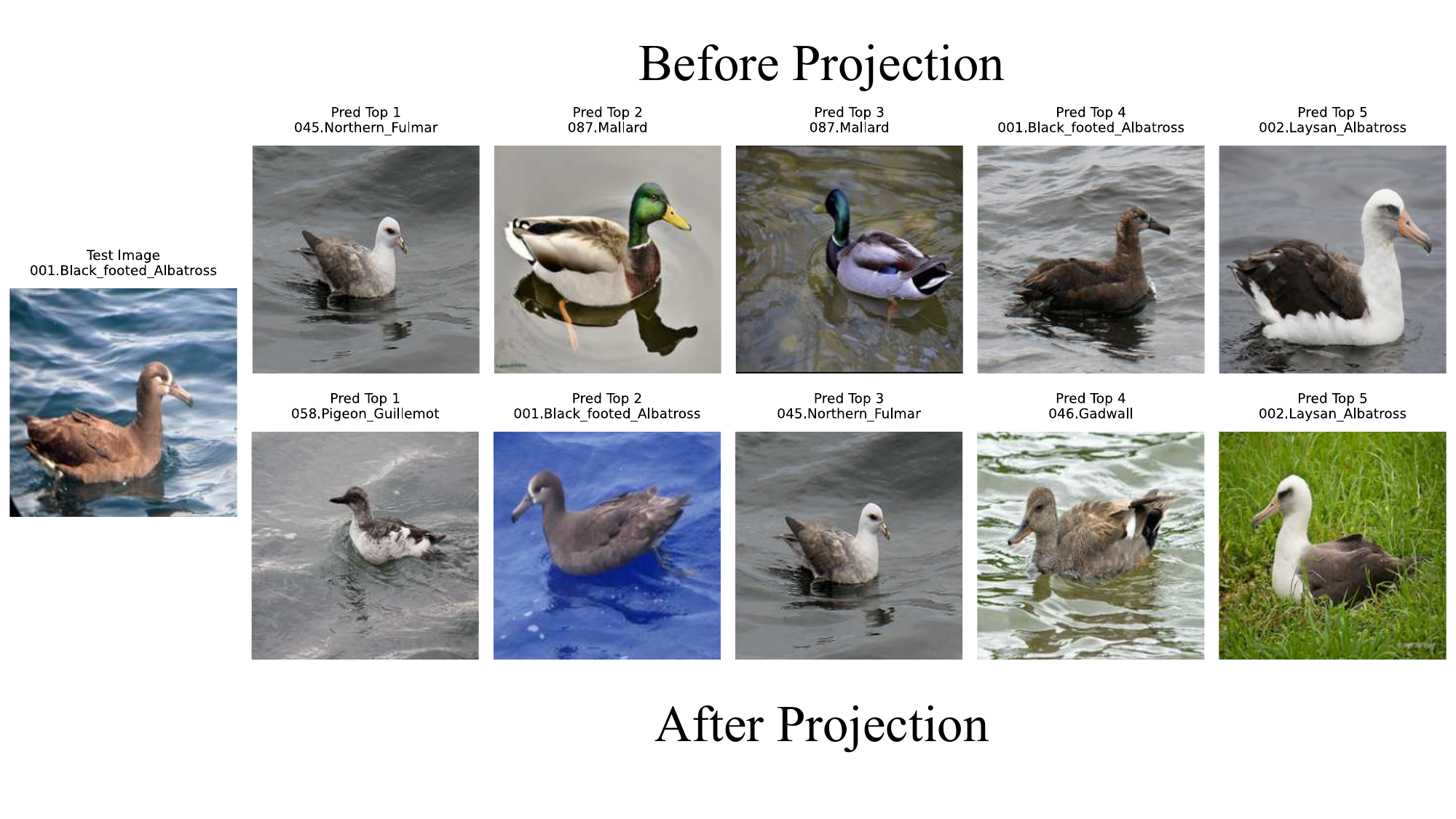}
            \caption{Top five retrieved images of Idefics2 image embeddings}
        \end{subfigure}

        \begin{subfigure}[b]{\columnwidth}
            \centering
            \includegraphics[width=0.8\columnwidth]{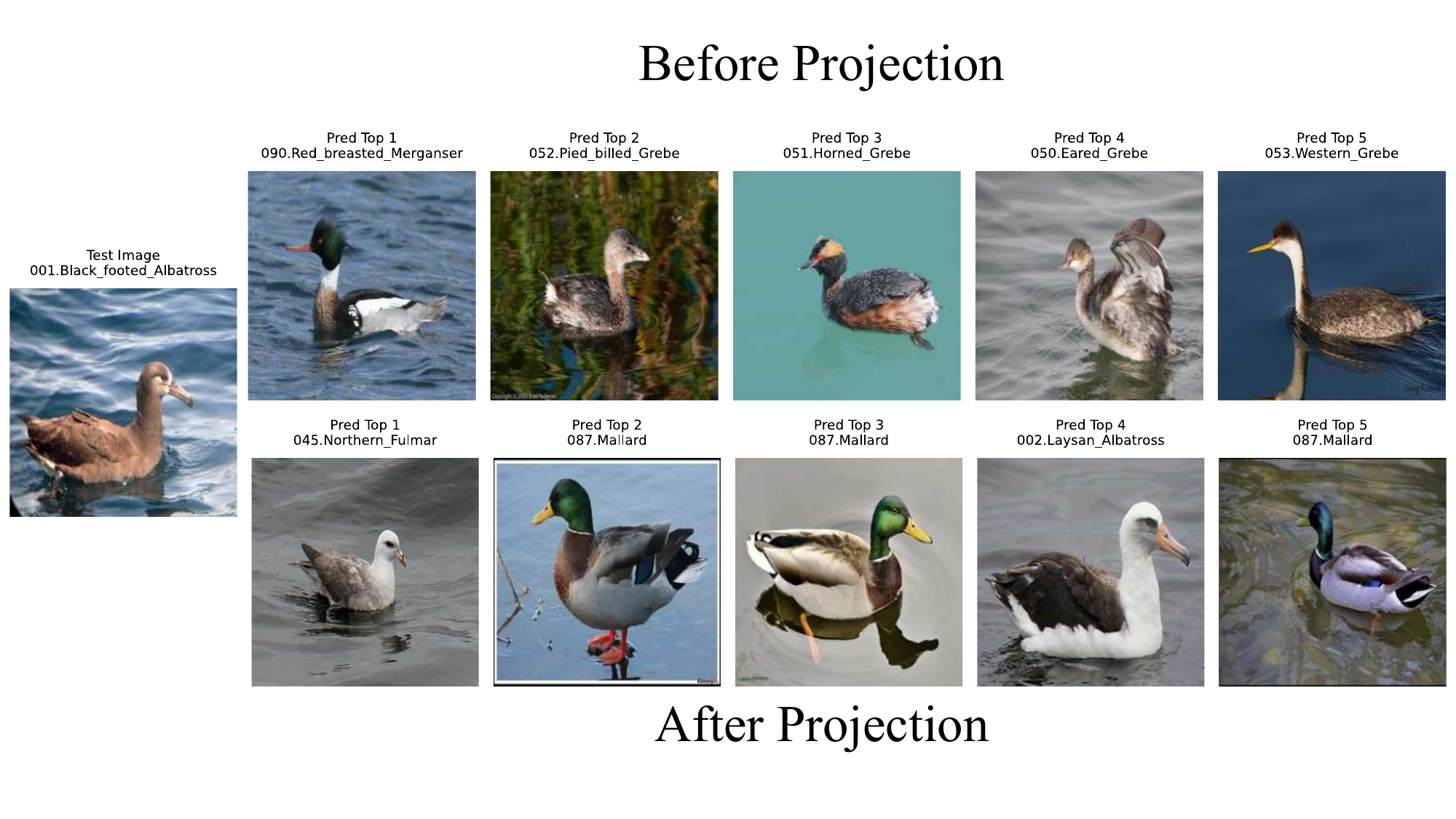}
            \caption{Top five retrieved images of Qwen2.5-VL image embeddings}
        \end{subfigure}
    
        \caption{Comparison of top five retrieved images of pre-projection (top) and post-projection (bottom) embeddings using different models on CUB test sample. Zero-shot retrieval based on fine-grained visual details is hard for all tested models.}
        \label{fig:cub}
    \end{figure}

In Figure~\ref{fig:knns}, we present more $k$-NN examples on comparison of searching with pre-projection (top) v.s. post-projection (bottom) embeddings. In Figure~\ref{fig:cub}, we present CUB image retrieval visualization with pre- and post-projection embeddings.

\subsection{Visualization of reconstruction loss and captioning performance}

\begin{figure*}[thb]
    \centering
    \includegraphics[width=\linewidth, trim=1.1em 0 0 0, clip]{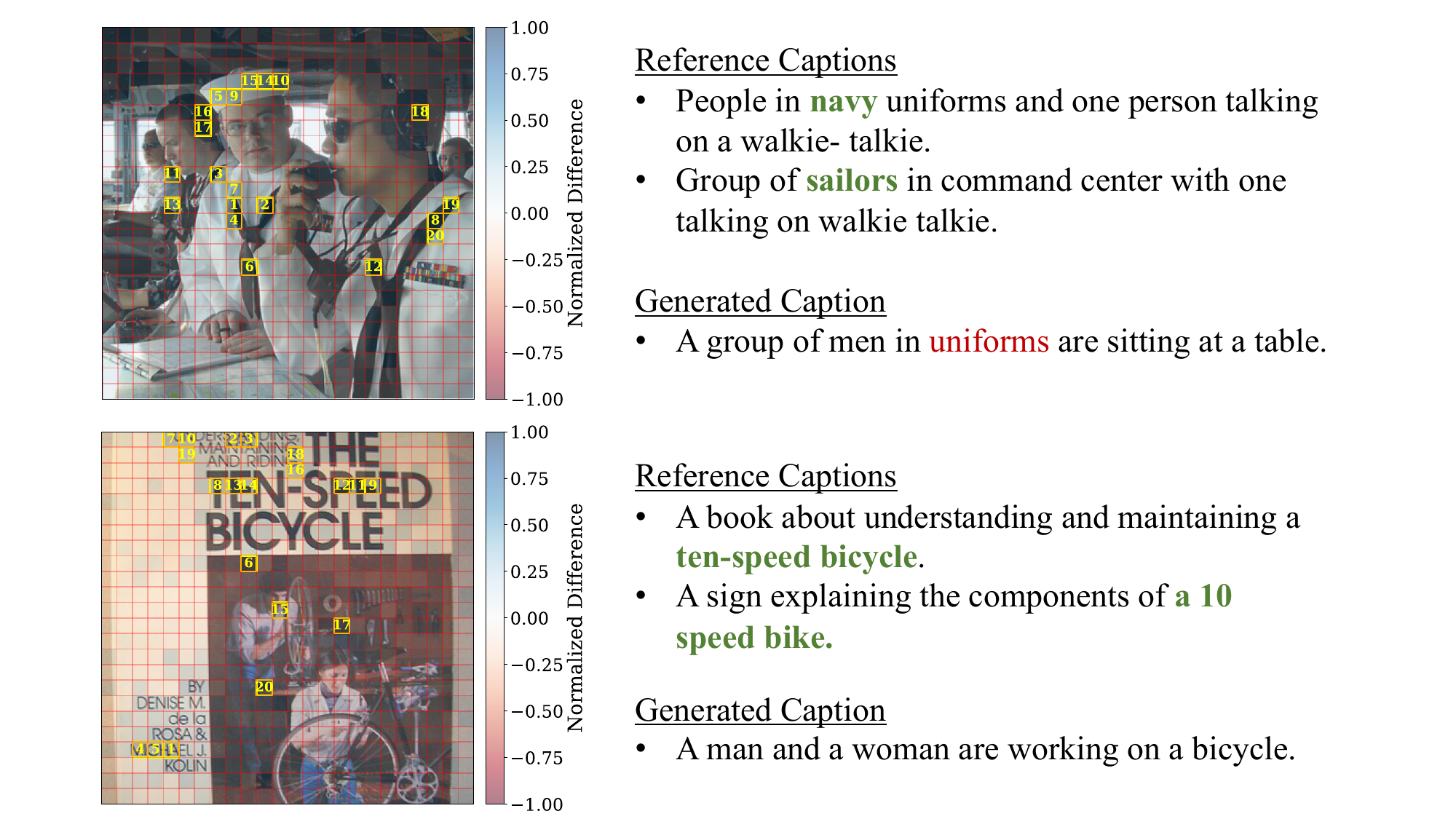}
    \includegraphics[width=\linewidth, trim=0 0 0.7em 0, clip]{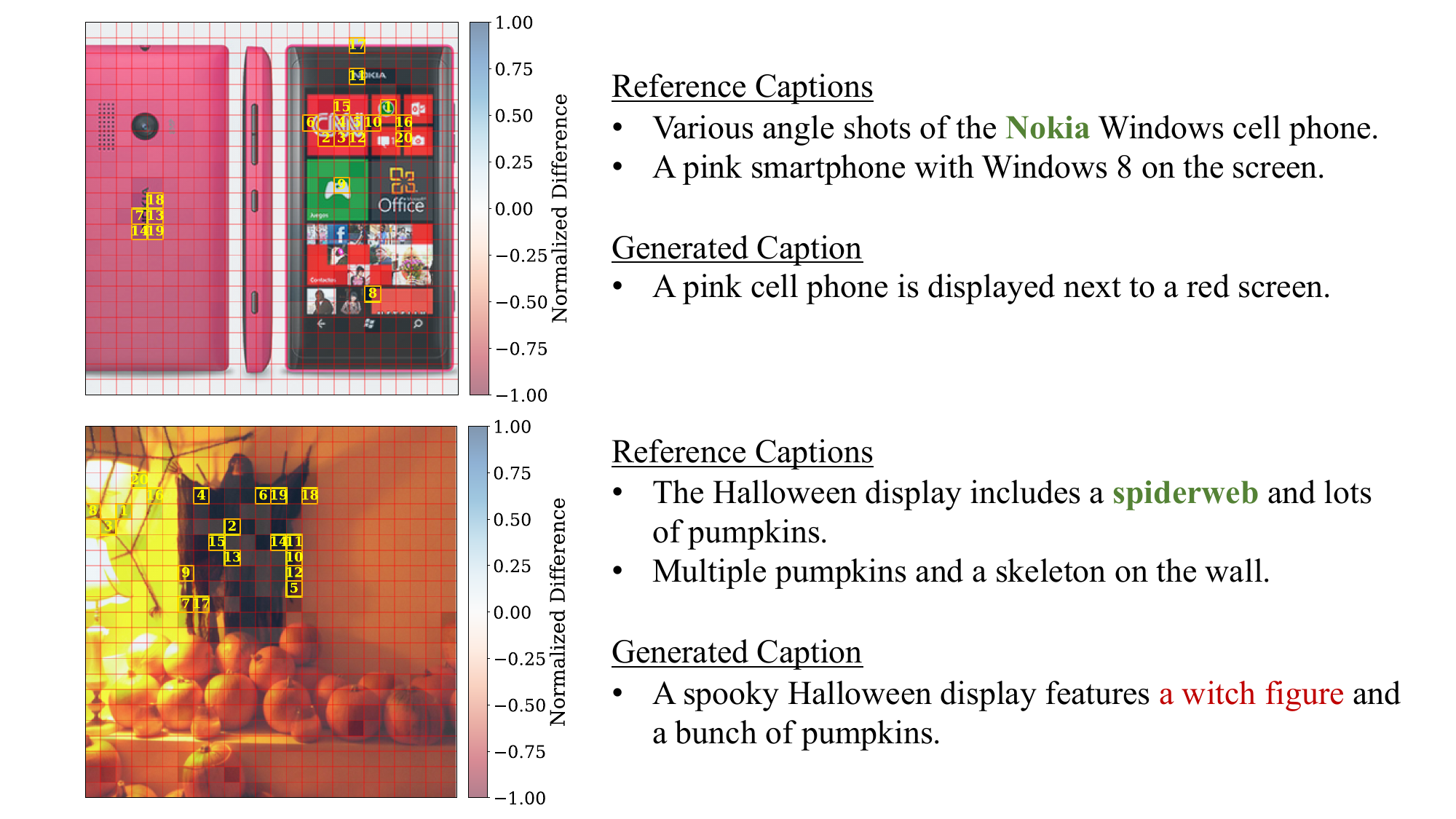}
    \caption{Visualization of low CIDEr score captioning samples and the reconstruction loss overlay with the input image. We can observe that details regarding the high loss patches are missing from the generated captions. High loss patches are marked in yellow squares.}
    \label{fig:cap-loss}
\end{figure*}

\section{Image Reconstruction with Different Embeddings}
\label{app:image_reconstruct}

Beyond neighbor-overlapping and embedding reconstruction, we aim to investigate how information loss manifests in the reconstructed images themselves. 
To explore this, we project different representations of visual features onto the input embedding space of a powerful image decoder to assess their reconstruction quality. 
However, image reconstruction performance depends on various factors, including the expressiveness of the image decoder. As such, this section serves as a preliminary exploration, and we encourage future work in this direction.

For our experiments, we use a fine-tuned VAE decoder\footnote{\url{https://huggingface.co/stabilityai/sd-vae-ft-mse}}, trained on the original VAE checkpoint from Stable Diffusion, trying to alleviate the influence of the decoder as a limiting factor in reconstruction quality. 
To align the sequence length between the vision encoder in the VLM and the expected input length of the VAE decoder, we employ a 6-layer Transformer encoder-decoder module with 4 attention heads. 
We train the aligner module on the COCO 2017 training set for 100 epochs with three objectives:
1) Embedding loss minimizing the difference between the VAE encoder embeddings and the aligned embeddings from the VLM’s visual encoder;
2) Reconstruction loss measuring the mean squared error (MSE) between the original and reconstructed images;
3) Latent loss quantifying the divergence between the mean and variance of the Gaussian distribution for diffusion.

\begin{figure}[ht]
    \centering
    \includegraphics[width=0.7\columnwidth]
    {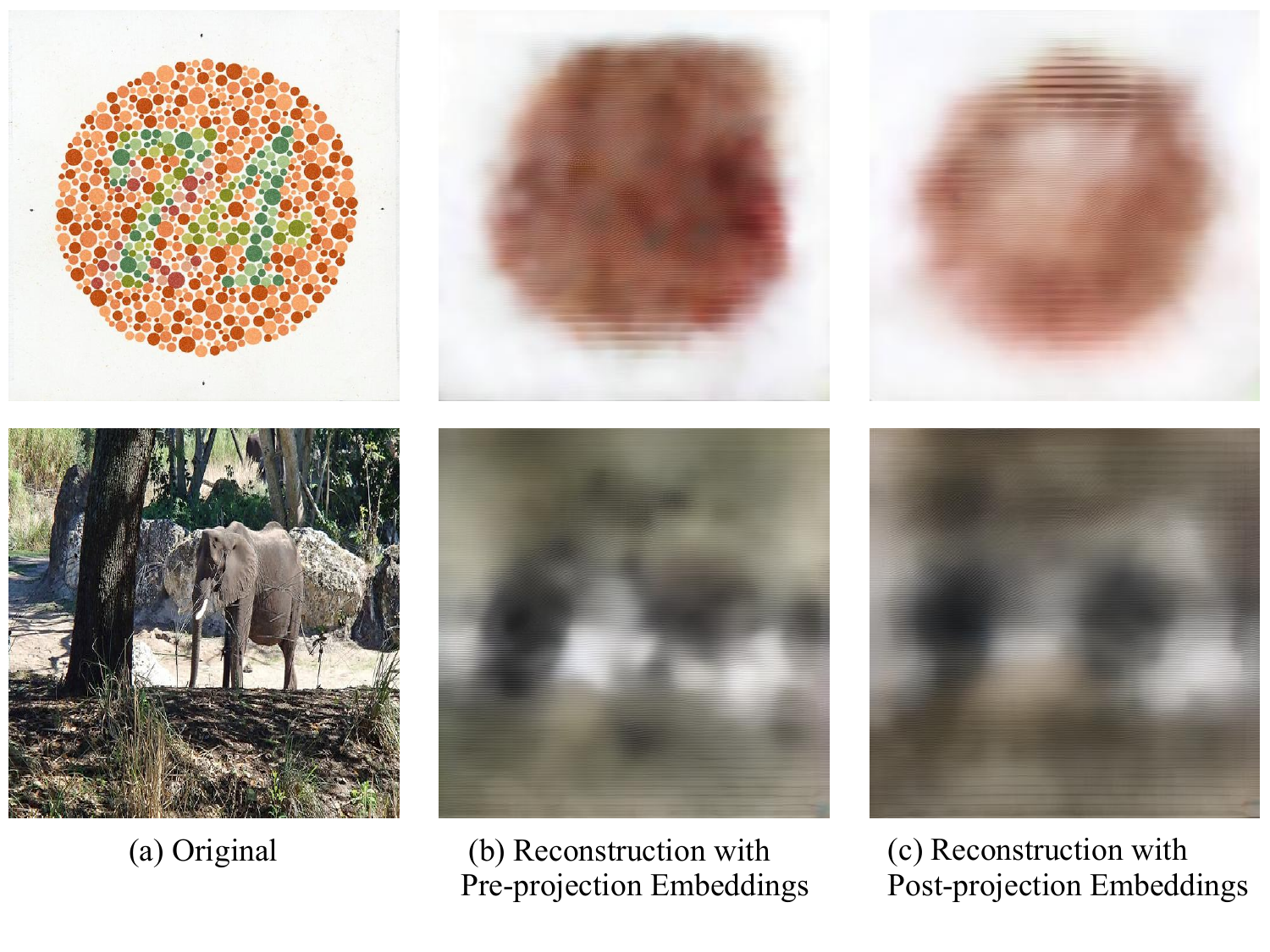}
    \caption{Image reconstruction with LLaVA pre-and post-projection embeddings on out-of-distribution (top) and in-distribution (bottom) examples. }
    \label{appfig:image_reconstruct}
\end{figure}

For the VLM, we use the LLaVA model in our experiments. 
We evaluate reconstruction performance on both an in-distribution image from the COCO 2017 dev split and an out-of-distribution image, as shown in Figure \ref{appfig:image_reconstruct}. 
When using embeddings before projection, the overall pixel-wise MSE reconstruction loss is $0.2128$, compared to $0.2443$ after projection. 
Figure \ref{appfig:image_reconstruct} illustrates the reconstructed images for both cases, where pre-projection embeddings yield similar contour preservation with post-projection embeddings.
\clearpage

\end{document}